\documentclass[sigconf]{acmart}
\usepackage{balance}
\def\BibTeX{{\rm B\kern-.05em{\sc i\kern-.025em b}\kern-.08emT\kern-.1667em\lower.7ex\hbox{E}\kern-.125emX}}
    
\copyrightyear{2019} 
\acmYear{2019} 
\setcopyright{acmlicensed}
\acmConference[KDD '19]{The 25th ACM SIGKDD Conference on Knowledge Discovery and Data Mining}{August 4--8, 2019}{Anchorage, AK, USA}
\acmBooktitle{The 25th ACM SIGKDD Conference on Knowledge Discovery and Data Mining (KDD '19), August 4--8, 2019, Anchorage, AK, USA}
\acmPrice{15.00}
\acmDOI{10.1145/3292500.3330782}
\acmISBN{978-1-4503-6201-6/19/08}

\begin{document}

% The "title" command has an optional parameter, allowing the author to define a "short title" to be used in page headers.
\title{AiAds: Automated and Intelligent Advertising System for Sponsored Search}

% The "author" command and its associated commands are used to define the authors and their affiliations.
% Of note is the shared affiliation of the first two authors, and the "authornote" and "authornotemark" commands
% used to denote shared contribution to the research.
\author{Xiao Yang*, Daren Sun, Ruiwei Zhu, Tao Deng, Zhi Guo, Jiao Ding, Shouke Qin, Zongyao Ding, Yanfeng Zhu}

\affiliation{%
  \institution{Baidu Inc.}
}
\email{{yangxiao04, sundaren, zhuruiwei, dengtao02, guozhi, dingjiao, qinshouke, dingzongyao, zhuyanfeng}@baidu.com}

%\author{Lars Th{\o}rv{\"a}ld}
%\affiliation{%
%  \institution{The Th{\o}rv{\"a}ld Group}
%  \streetaddress{1 Th{\o}rv{\"a}ld Circle}
%\email{larst@affiliation.org}
%\affiliation{%
%  \institution{Inria Paris-Rocquencourt}
%\affiliation{%
% \institution{Rajiv Gandhi University}
%\affiliation{%
%  \institution{Tsinghua University}
%  \streetaddress{30 Shuangqing Rd}
%\affiliation{%
%  \institution{Palmer Research Laboratories}
%  \streetaddress{8600 Datapoint Drive}
%\affiliation{\institution{The Th{\o}rv{\"a}ld Group}}
%\email{jsmith@affiliation.org}
%\affiliation{\institution{The Kumquat Consortium}}
%\email{jpkumquat@consortium.net}

%
% By default, the full list of authors will be used in the page headers. Often, this list is too long, and will overlap
% other information printed in the page headers. This command allows the author to define a more concise list
% of authors' names for this purpose.
\renewcommand{\shortauthors}{Yang, et al.}

%
% The abstract is a short summary of the work to be presented in the article.
\begin{abstract}
Sponsored search has more than 20 years of history, and it has been proven to be a successful business model for online advertising. Based on the pay-per-click pricing model and the keyword targeting technology, the sponsored system runs online auctions to determine the allocations and prices of search advertisements. In the traditional setting, advertisers should manually create lots of ad creatives and bid on some relevant keywords to target their audience. Due to the huge amount of search traffic and a wide variety of ad creations, the limits of manual optimizations from advertisers become the main bottleneck for improving the efficiency of this market. Moreover, as many emerging advertising forms and supplies are growing, it's crucial for sponsored search platform to pay more attention to the ROI metrics of ads for getting the marketing budgets of advertisers.  

In this paper, we present the AiAds system developed at Baidu, which use machine learning techniques to build an automated and intelligent advertising system. 
By designing and implementing the automated bidding strategy, the intelligent targeting and the intelligent creation models, the AiAds system can transform the manual optimizations into multiple automated tasks and optimize these tasks in advanced methods. AiAds is a brand-new architecture of sponsored search system which changes the bidding language and allocation mechanism, breaks the limit of keyword targeting with end-to-end ad retrieval framework and provides global optimization of ad creation. This system can increase the advertiser's campaign performance, the user experience and the revenue of the advertising platform simultaneously and significantly. We present the overall architecture and modeling techniques for each module of the system and share our lessons learned in solving several key challenges. Finally, online A/B test and long-term grouping experiment demonstrate the advancement and effectiveness of this system.
\end{abstract}



%
% Keywords. The author(s) should pick words that accurately describe the work being
% presented. Separate the keywords with commas.
\keywords{sponsored search, automated bidding, intelligent targeting, intelligent creation}

%
% A "teaser" image appears between the author and affiliation information and the body 
% of the document, and typically spans the page. 
%  \Description{Enjoying the baseball game from the third-base seats. Ichiro Suzuki preparing to bat.}

%
% This command processes the author and affiliation and title information and builds
% the first part of the formatted document.
\maketitle

\section{Introduction}
Sponsored search is an indispensable part of the business model in the modern online advertising market. According to the Statista \cite{Statista19} , revenue in the search advertising segment amounts to 96 billion dollars in 2018. By determining the keyword searches that are most relevant to their business's offerings, advertisers create ads and bid on relevant keywords to place their ads in the search results. The display and position of the ads are determined by a real-time auction when users are searching for corresponding terms. Sponsored search provides considerable revenue for general search engine services such as Google and Baidu. It is a huge online market in which tens of billions of auctions are held every day and several different types of business offerings are distributed to various users. 

Traditionally, the pay-per-click pricing model and the keyword targeting technology are the two keys to the business success of sponsored search. The pay-per-click model is the most common payment method, in which an advertiser pays a publisher only when the ad is clicked. Keyword targeting provides an accurate match between the search query and the advertising terms, in which advertisers bid on keywords that are related to their products or services to display their ads to the targeted audience on the search result pages. After more than 20 years, the traditional model of sponsored search gradually shows its shortcomings and limits. 

Firstly, keyword targeting requires that the advertiser should select plentiful keywords relevant to their business to increase the coverage of related and targeted search traffic. As users can express their search intent in a variety of different queries, it is challenging for an advertiser to find all the terms relevant to their offer from this huge inventory of possible terms. Due to the limit of exploration capability and knowledge about the broad scope of different keywords, most advertisers can only bid on a handful of relevant keywords which lead to insufficient advertising effect. Though the sponsored system has provided some keyword recommendation tools and multiple match types such as exact, phrase and broad and so on, but the keywords and match type set by advertisers still play an important role in ad retrieval, and all the retrieved ads should be subject to these keywords. An ad will not be retrieved even if search queries are related if the advertiser does not bid on corresponding keywords or set the match type correctly. As the biggest search engine in China, Baidu has billions of search queries and subsequent page views each day and the new queries grow dramatically. Massive search traffic and highly dynamic search intents pose huge challenges for advertisers in their manual selection of keywords. In this situation, the keywords selected manually by advertisers determine the upper-bound of commercial search volumes, sponsored search engines cannot achieve the actual global optimal matching between search queries and ads, the advertiser also cannot get sufficient ad impressions if he doesn't have an expert for selecting keywords. 

Secondly, in the pay-per-click pricing model, since the keyword bids play a direct role in the ranking and pricing of ads, advertisers should bid on each keyword cautiously. Due to the trade-off between usability and pertinency, most of the sponsored search systems adopt the keyword-level bidding language, and advertiser sets the bid price for keyword to represent his/her willingness to pay based on the click value of this keyword. To keep and improve campaign performance, advertiser must adjust the bid prices frequently according to the objectives (KPIs) and various bidding feedback signals such as cost-per-click, ad impressions or clicks, budget data, and so on. As sponsored search market has a highly dynamic and competitive environment, it takes a long time to build a stable manual bidding strategy. The aforementioned keywords selection problem also brings more workloads to the bidding optimization task, when the advertiser selects more keywords, then more efforts need to be to put into the bidding strategy. Advertisers from big businesses usually recruit search engine marketing(SEM) experts to optimize their campaigns, while maintaining plenty of keywords and their bidding optimization is non-trivial especially for small and local businesses that don’t have dedicated marketing staff. Moreover, keyword-level bidding language is too coarse-grained to represent the real value of each search traffic, it only can give an average click value for a cluster of the search volume. Even for the same search keyword, different time or different locations or different users may generate different click values for advertisers. The desired bidding language should support adjusting bid prices based on where, when, and how people search, but it will bring the curse of dimensionality to the manual bidding optimization.
   
In the meantime, though the global digital advertising market is still growing, the percentage of digital ad revenue captured by search is falling. There are many emerging advertising forms such as social media advertising, video advertising, contextual or native advertising and so on. As these new digital media have massive active users and can provide novel advertising products, the search engine is no longer the dominated traffic source for online advertising. For the performance advertisers, they optimize for the sales, signups, or other so-called conversions generated directly from their ads, and they will choose to allocate their budget to multiple advertising media based on the return on investment (ROI) metric. In this competitive situation, the ROI or the cost-per-acquisition(CPA) of advertising becomes more and more important to digital media to attract advertisers. For the sponsored search market, the traditional metrics such as cost-per-click, ad clicks are very indirect for performance advertisers to control and optimize their conversions and ROI, and the keyword-level bidding optimization is inefficient for increasing the advertising performance.

Thirdly, the creation of search advertising plays an important role in ad performance, especially impacts the click-through rate(CTR) of ad. Traditional format of search advertising is the textual ad which has three creation parts: a headline text or title, a display URL, and a description text. Advertisers can optimize their ad creations by using more attractive title or promotional description. Recently, as online ad offerings become increasingly complex, rich ads with features such as larger formats, reviews, maps, sitelink extensions, call or app buttons, images, and many other decorations that result in an advertiser having several possible ads with varying sizes and layouts. When the content and layout for ad creation become more diverse and rich, how to design better creations faces a larger combination and optimization space for advertisers. The sponsored search system provides many ad formats, advertisers should prepare relevant materials and select related formats to set the specific ad creations. Due to multiple content configurations and layouts available to advertisers, the manual selection or optimization of ad creations is very hard to advertisers. 

To address these main challenges in sponsored search, we propose and build an automated and intelligent advertising system, called AiAds. As the above three problems are all the bottlenecks of manual optimization for advertisers, we use machine learning models to solve these problems by transforming the traditional manual targeting, bidding, ad creation tasks into automated tasks. Base on this system, automated bidding, intelligent targeting and intelligent creation are integrated to support a more intelligent advertising system, and the advertiser can entrust the performance optimization of ads to this system by only setting their target. In this paper, we focus on sharing our experience in building the AiAds system and report empirical results after deployed it in Baidu. The main contributions of this work are as follows:
\begin{itemize}
\item We present a straightforward bidding language and corresponding automated bidding strategy for advertisers to optimize their campaign performance directly. We show the basic data requirement and model architecture used in bidding strategy. The new bidding language and strategy also extend the traditional pay-per-click pricing model and bring new challenges in designing the auction mechanism. 
\item Based on the new bidding language, we break the limits of traditional keyword targeting method. By using the more straightforward retrieval and matching model, the system can optimize a more optimal matching and selection between search queries and ads in an end-to-end manner.  
\item We present a componentized framework for designing and generating ad creations which can use the materials to optimize the content and layout of advertising automatically.
\item We conduct the online A/B test and long-term grouping experiment on live traffic of Baidu, all results show that the AiAds system can significantly increase the advertising performance, increase the revenue of the search engine and increase the user experience. 
\end{itemize}
  
The rest of the paper is organized as follows. We start by introducing the related works in Section 2. We then introduce the overall systems architecture in Section 3. The bidding language and automated bidding strategy are described in Section 4. In Section 5, we talk about the models used to deploy our solutions for intelligent targeting. The intelligent creation framework is described in Section 6. The experimental results are discussed in Section 7. We finally conclude this paper in Section 8.

\section{Related work}
As the sponsored search is a hot research direction, there has been much work and an abundant literature on the optimization of different aspects \cite{qin2015sponsored,feldman2008algorithmic} of the sponsored search systems. 

For the keyword targeting and ad retrieval task, a vast amount of methods have been proposed. For example, how to improve the broad match of bid keywords for a given query was studied in\cite{broder2008search,broder2009online,even2009bid,zhang2007query}.  An adaptive algorithm was proposed\cite{gupta2009catching} which could utilize arbitrary similarity functions and catch the dynamics in the broad match. Generating bid keywords for some given landing pages of the advertisers was discussed in\cite{ravi2010automatic}. The advertisability of tail queries in sponsored search system was studied in\cite{pandey2010estimating}. In addition, some works have been proposed for bid keyword recommendation. A novel algorithm for advertising keywords recommendation was presented in\cite{zhang2012advertising} By leveraging the contents of Wikipedia. Recommending a group of relevant yet less-competitive keywords to an advertiser was proposed in\cite{zhang2014bid}. Most of these works have considered keyword relevance as a key factor in their algorithms. However, these methods can not overcome the limitation of keyword-based ad retrieval. In order to enhance sponsored search ad retrieval, a number of extractive summarization techniques for landing pages were explored in\cite{choi2010using}. A query-ad semantic matching approach based on embeddings of queries and ads was proposed in\cite{grbovic2016scalable}, and the embeddings were learned on user search session data in an unsupervised manner. Machine translation model\cite{song2017translation} was used to translate a natural language query into a keyword. A collaborative filtering algorithm based on the bipartite graph was presented in\cite{anastasakos2009collaborative}. Similarly, a network-based\cite{yan2017beyond} ad retrieval framework was proposed. For our intelligent ad retrieval task, we use some models about mining and learning for heterogeneous networks\cite{dong2017metapath2vec}, and leverage deep learning techniques for the semantic matching problem\cite{shen2014learning,shen2014latent} in the information retrieval and recommendation systems.  
     
Bidding optimization has been well studied in sponsored search. A systematic exploration of a natural class of greedy bidding strategies was undertaken in\cite{cary2007greedy}. Bid optimization and generation for advanced match was studied in\cite{even2009bid,broder2011bid}. The joint optimization of campaign budget allocation and bid price setting was proposed in\cite{zhang2012joint}. However, most previous works focused on the keyword-level auction paradigm, and the conversion or ROI metrics was not taken into the bidding strategy. A solution to automatically adjust the bid price for advanced matching based on conversion rate prediction was presented in\cite{rey2010conversion}. Similarly, a bid optimizing strategy called optimized cost per click was proposed in\cite{zhu2017optimized}. A reinforcement learning based real-time bidding strategy for sponsored search was proposed in\cite{zhao2018deep}. In the industry of sponsored search, there are also some bidding strategies tools\cite{googlebid19,bingbid19}, such as Enhanced CPC, Target CPA, Maximize Conversions, and so on. The conversion-based bid strategies are more suitable for the performance advertisers to optimize their ROI. 

For the rich advertisements in sponsored search, there are a lot of ad extensions or formats in Google\cite{Google19} and Baidu\cite{Baidu19}. The ad creation optimization was less studied by academia. But there was some research work about how to design the combinatorial auction mechanism for rich ads \cite{cavallo2017sponsored,hartline2018fast,bachrach2014optimising}. Two sampled rich search ads in Baidu are presented in Figure 1, there are text creations, images, download links, phone call button, sitelink extensions in the ads. 

\begin{figure}[h]
  \centering
  \includegraphics[width=\linewidth]{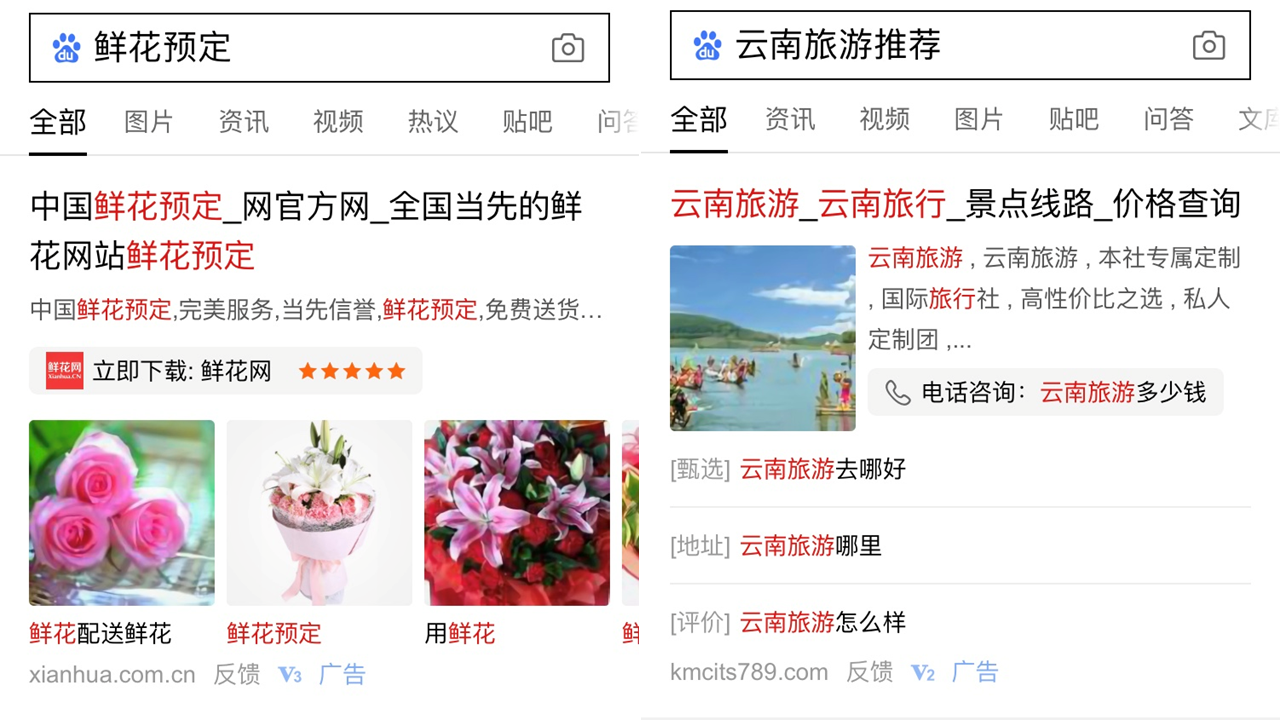}
  \caption{The formats of ads in Baidu.}
  \Description{The ad formats of Baidu.}
\end{figure}

\section{Systems Architecture}

Before diving into how the system is built, let us first introduce the overall system architecture of AiAds. As showed in Figure 2, the whole system is composed of five modules: the unified data center, the basic models of search ads, the intelligent targeting service, the automated bidding engine, and the intelligent creation generator. The unified data center provides the basic data requirement for the downstream tasks and consists of several data sources such as ad click data, ad conversion data and search log data, and so on. The basic CTR and CVR model are built to model and predict the click-through rate and conversion rate for each ad. Some ad relevance models are built to quantize the relevance between queries and ads, and to optimize the user experience. When users submit search requests, the intelligent targeting service retrieves all the related ads directly, and the automated bidding engine produces bid prices for each ad based on the real-time models and the target of advertisers. Finally, based on the candidate materials and contents, the intelligent creation generator makes the combination and the generation of ad creations to display.  

\begin{figure*}[h]
  \centering
  \includegraphics[width=12cm,height=5cm]{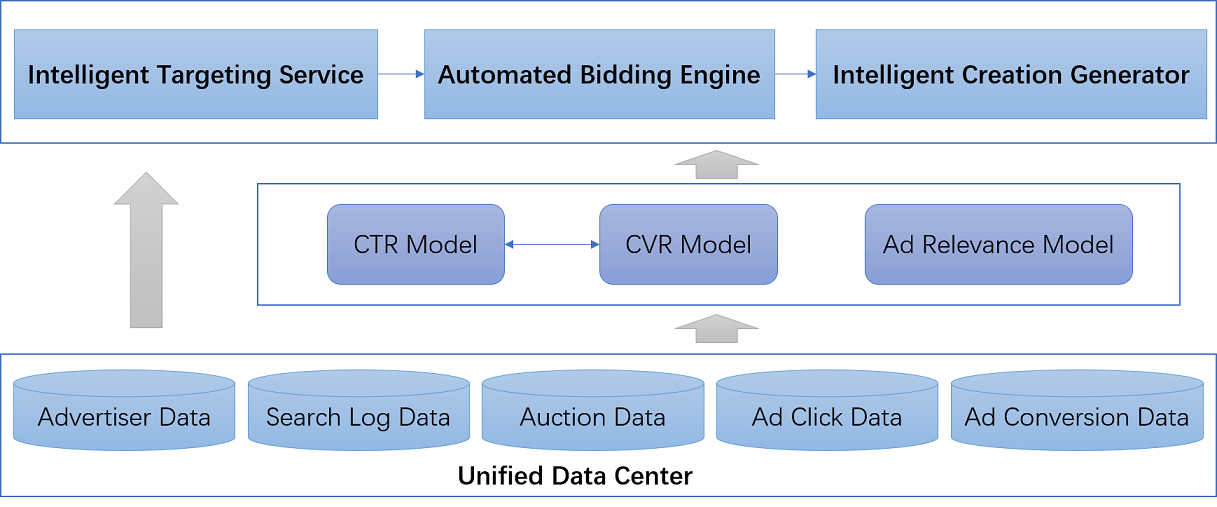}
  \caption{The Systems Architecture of AiAds.}
  \Description{The Systems Architecture of AiAds.}
\end{figure*}

\subsection{Basic Setting}

We now describe the setting and notation we use throughout the paper. Our setting is a standard sponsored search. First, we introduce some notation. Assume that there is a set \(N = \{1, ..., n\}\)of \(n\) advertisers. For each advertiser $i$, he selects $K_i$ related keywords and the $ k_{ij}$ indicates the $j$th keyword selected by advertiser $i$, and he creates $A_i$ ads and the $ a_{ij}$ indicates the $j$th ad created by advertiser $i$. As already mentioned, the Pay-Per-Click pricing model is employed. For each $ k_{ij}$, advertiser $i$  has a private click value $v_i(k_j)$, expresses the maximum price per click he is willing to pay. In order to participate in the auction, advertiser $i$ is required to submit a bid $b_i(k_j)$. $b_i(k_j)$ is a proxy of the value $v_i(k_j)$ but might not exactly equal $v_i(k_j)$ due to the strategic behaviors of the bidders. Submitting a bid $b_i(k_j)$ guarantees to advertiser $i$ that he will not be charged a price higher than $b_i(k_j)$ per click. The vector of the advertiser $i$'s bids, $b_i$, will usually be referred to as his bidding strategy profile. When the displayed ad of advertiser $i$ associated with $ k_{ij}$ is clicked, he will be charged a price $p_i(k_j)$, which is also called cost-per-click(CPC), and we have $p_i(k_j) \leq b_i(k_j)$.     

CTR and CVR are two important parameters in the context of sponsored search. The click-through-rate(CTR)is the probability that a given ad will be clicked when displayed, and the conversion rate(CVR) is the probability that a ad conversion(A conversion is an action that's counted when someone interacts with advertiser's ad and then takes an action that defined as valuable to his business such as purchased a product, signed up for newsletter, called advertiser's business, or downloaded app and so on) be acquired after click this ad. So we can have that $CTR = \frac{clicks}{impressions}$ and $CVR = \frac{conversions}{clicks}$. We use $ctr(a_{ij})$ and $cvr(a_{ij})$ to represent the actual CTR and CVR of the ad $ a_{ij}$, and the $pctr(a_{ij})$ and $pcvr(a_{ij})$ to represent the predicted CTR and CVR. Another important quantity is the cost-per-acquisition (CPA) which represents the average cost for each conversion, so we have $CPA = \frac{cost}{conversions}$.

Advertisers can be split into two categories: brand and performance. Brand advertisers aim for long-term growth and awareness, they have a mandate to meet a specific
business goal—showing impressions to an audience, generating clicks, or maximizing revenue—driven by long-term considerations instead of immediate profit. Performance advertisers optimize the immediate tradeoff between value—measured as sales, sign-ups, or other so-called conversions generated directly from their ads—and cost. Return on investment(ROI) has been the standard metric for measuring this tradeoff across all types of advertising for decades. ROI measures the ratio of the profit obtained (“return”) to the cost or price paid (“investment”), i.e., the density of profit in cost: $ROI = \frac{Revenue-Cost}{Cost}$. Being a density metric, unconstrained maximization of ROI is not sensible, instead, performance advertisers come with an ROI constraint and maximize their revenue. 

The performance advertisers are the main customers for sponsored search and many other digital advertising platforms. To prevent advertisers shifting ad budgets, advertising platforms should keep a competitive ROI metric. 

\section{Automated Bidding Engine}
In this section, we will cover the details of how we design and build the automated bidding engine to address the problems of traditional keyword-level manual bidding optimization. 

\subsection{The Bidding Language}

In sponsored search, advertisers set and optimize their bids to achieve a specific goal for their business. As plenty of keywords should be maintained and highly dynamic and competitive auction environment, the keyword-level bidding language is very inefficient and brings great challenges to manual bidding optimization. Nowadays, for the performance advertisers, their goal is to maximize their profit with an ROI constraint. We can formulate the utility function of performance advertisers as:
\begin{equation}
\begin{aligned}
  U_i &= (revenue - cost) \\ &= (sale\ value * sales - conversions * CPA) \\ &= (sale\ value * sale\ rate -  CPA) * conversions \\& s.t. \   ROI_i \geq \gamma_i
\end{aligned}                       
\end{equation}
And the $sale\ rate = \frac{sales}{conversions}$. For the ROI constraint, we can have:
\begin{equation}
\begin{aligned}
  ROI_i \geq \gamma_i &\Rightarrow \frac{revenue-cost}{cost} \geq \gamma_i \\ &\Rightarrow \frac{sale\ value * sale\ rate -  CPA}{CPA} \geq \gamma_i \\ &\Rightarrow  CPA \leq \frac{sale\ value * sale\ rate}{1+\gamma_i} 
\end{aligned}                       
\end{equation}
From the equation (2), the ROI constraint can be transformed into a target CPA constraint. So the utility function of performance advertisers is their profit subject to a target CPA. 

In the practice, as the digital advertising market has more and more supplies such as social media and video media and both these media have ample traffic volumes, so the performance advertisers face the budget allocation problems. Rational advertisers will allocate their budgets based on the ROI of supply media, and the media with higher ROI and more conversions will get more budgets from the advertisers.

Therefore, for the real optimization goals of performance advertisers, the keyword-level click value based bidding language is too indirect and coarse-grained, and advertisers should make lots of efforts in accumulation of clicks and calculation of CVR. To tackle this problem, we provide a straightforward target CPA based bidding language and advertiser can set the target CPA at campaign or ad group level directly, to replace the manual keyword-level bid. The target CPA is a straightforward representation of advertiser's ROI constraint, and the advertising platform should provide as many conversions as possible for the advertiser under this constraint. 

\subsection{Automated Bidding Strategy}

Based on this new bidding language, advertisers only need to set and optimize their global target CPA for conversions, and the concrete real-time bidding for each auction can be done by the automated system. 
From the equation (1) and equation (2), we can conclude the optimizing objective of the advertiser, which is:
\begin{equation}
\begin{aligned}
  &max \{conversions\} \\ & s.t. \   CPA \leq target \ CPA
\end{aligned}                       
\end{equation}

The revenue of sponsored search system can be formulated as:
\begin{equation}
\begin{aligned}
  Revenue &= \sum_{i=1}^{M} {click_i * CPC_i} \\ &= \sum_{j=1}^{N}  {conversion_j * CPA_j}
\end{aligned}                       
\end{equation}

From the equation (4) and equation (3), we can conclude the optimizing objective of the automated bidding strategy, which is:
\begin{equation}
\begin{aligned}
  &max \{conversions\} \\ & s.t. \   min|CPA - target \ CPA|
\end{aligned}                       
\end{equation}
And the auction mechanism should ensure the IC(incentive compatibility) for expressing the target CPA, under the setting with multiple sellers and multiple bidders. 

The equation (5) is also can be regarded as the optimizing objective of the AiAds system, and we can find that this objective is all-win to advertisers, sponsored search platform and users. 

To optimize the equation (5), we propose a multiplicative automated bidding strategy to produce the real-time bidding (RTB) for each auction by combining multiple real-time bidding factor(also can be solved as a feedback control problem), which is:
\begin{equation}
\begin{aligned}
  RTB = CPA * pcvr * AF * BF * CF * Alpha
\end{aligned}                       
\end{equation}
The objective of this bidding strategy is to minimize the CPA gap and to maximize the conversions. And each bidding factor is described as follows:
\begin{itemize}
\item{\textbf{CPA}}: The target CPA set by the advertiser.
\item{\textbf{pcvr}}: The CVR of ad predicted by machine learning model.
\item{\textbf{AF}}: The auction factor which is used to quantify the statistical gap between bid and CPC. As the bidding strategy is to optimize the conversions and make the average CPA close to target CPA, the statistical gap information can be used to increase proximity to the target value. We use a gradient boosting framework\cite{ke2017lightgbm} to model the gap, the label is $\frac{bid}{CPC}$, and the features consist related statistics about the query, the advertiser and the auction context, other bidders' information cannot be used due to the incentive problem.    
\item{\textbf{BF}}: The budget factor which is used to spend budget smoothly over the time in order to reach a wider range of audience accessible throughout a day. A smart pacing approach\cite{agarwal2014budget,xu2015smart,lee2013real} is used to adjust the bid price based on the prior performance distribution in an adaptive manner by distributing the budget optimally across time. 
\item{\textbf{CF}}: The calibration factor which is used to make the predicted CVR more close to the real CVR. The optimized isotonic regression\cite{zadrozny2002transforming} model and binning method are used to calibrate the $pcvr$ values. 
\item{\textbf{Alpha}}: The alpha factor is a dynamic parameter generated by a reinforcement learning model. As the highly dynamic auction environment in a day, real-time feedback information must be used to boost the achievement of target CPA. We split the whole day into eight time buckets, and the optimization of daily achievement of target CPA can be modeled as an MDP. We design a general representation for states as $s=<tid, sr, pg, cg, cd>$, where $tid$ denotes the bucket number, $sr$ denotes the spent ratio of budget, $pg$ denotes the pcvr gap between the current bucket and the last one, $cg$ denotes the CPC gap between the current bucket and the last one, and $cd$ denotes the relative diff between current CPA and target CPA. The reward function $r=min\{15, \frac{target\ CPA}{|real\ CPA - target\ CPA|}\}$ is used, and the action space is 31 discrete numerical ratios from [-3, 3] such as -3, -2.8, ..., 2.8, 3. We adopt a DQN algorithm similar to\cite{mnih2013playing,mnih2015human} which employs a deep neural network with weights $\theta$ to approximate $Q$ value. By the Alpha factor, we can model the bidding strategy as a dynamic interactive and sequential control process in a complex  environment rather than an independent prediction or optimization process.
\end{itemize}

Based on this automated bidding strategy, fine-grained and auction-time bidding comes true in sponsored search, and abundant signals can be used in bidding optimization.  
\subsection{The CVR Model}

From the equation (6), the $pcvr$ play an important role in the automated bidding strategy. If the CVR models are only trained with samples of clicked impressions, the model will have poor generalization ability in plenty of un-clicked impressions such as new ads. To tackle this problem, we put the training of the CVR model and the training of CTR model together by using a multi‑task model architecture and shares the lookup table for the embeddings of common features. The overall model architecture is presented in Figure 3. Moreover, as the conversion types are diverse, the network of CVR model also adopts a multi‑task architecture.
\begin{figure}[h]
  \centering
  \includegraphics[width=\linewidth]{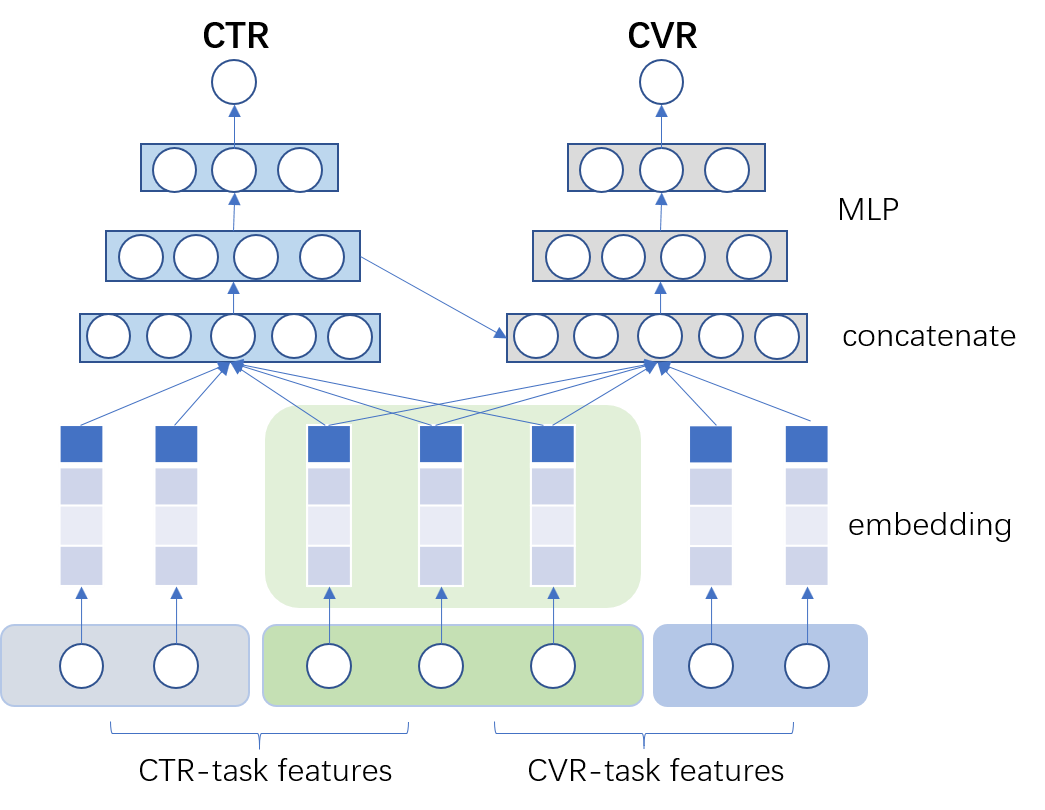}
  \caption{Architecture overview of CVR model.}
  \Description{Architecture overview of CVR model.}
\end{figure} 

\subsection{Auction Design for ROI-constrained Advertisiers}

The basic setting and assumption of traditional auction design in sponsored search(such as GSP, VCG, Myerson, etc.) are that:
\begin{itemize}
\item One seller, multiple bidders.
\item The utility function of bidder is quasilinear for single auction $U_i = (click\ value - CPC)*CTR$.
\item One-shot auction or the different auctions are independent.
\end{itemize}

In fact, since the utility function of performance advertisers is close to the equation (1), the traditional quasilinear utility function for single auction is no more hold. The new bidding language and bidding strategy also bring new challenges in designing the auction mechanism for sponsored search. 
Since in practice the ROI is computed in average over time, it creates a dependency between auction. Bidders are less sensitive to their obtained ROI in a single auction but are instead chiefly concerned about their expected ROI across many auctions.
The real condition of auction design for ROI-constrained advertisers is that:
\begin{itemize}
\item Multiple sellers, multiple bidders.
\item The utility function of the bidder is equation (1).
\item Repeated and sequential auctions, auctions are dependent and context-aware. 
\end{itemize}  

For the ROI-constrained bidder, the optimal mechanism design of ad auction should be aware of the ROI metric. There are some work show\cite{golrezaei2018auction,heymann2018roi,wilkens2016mechanism} that the different points in designing auctions for ROI-constrained bidders. 

For the performance advertisers, the two main changes in our auction mechanism are:
\begin{itemize}
\item According to\cite{golrezaei2018auction}, keyword-based reserve prices are removed, instead, personalized and CVR-based reserve prices are set.
\item To model and optimize the sequential auctions problem, we adopt the bank account mechanism\cite{mirrokni2017non,mirrokni2016dynamic} framework to increase the efficiency of the allocation and the revenue of sponsored search platform. We keep a state variable(balance) $b_t$ for each buyer based on his outcome of historical auctions, and use the $b_t$ in the next auction with the balance-independence property, dynamic reserves are also managed via bank accounts. 
\end{itemize}  
 
\section{Intelligent Targeting Model}

The traditional keyword targeting model restricts the range of ad retrieval, given a search query, the system must retrieve the related keywords, and then retrieve the ads related to these keywords. This two-stage retrieval procedure may lose many candidates, even for the advanced broad match type, the set of keywords selected by advertisers is still the bottleneck for ad retrieval. 

Is the keyword really necessary for sponsored search? Based on the high-level target CPA bidding language, the keyword-level bids are needless, so we can break the limits of keyword targeting method. With the help of automated bidding strategy, the keyword targeting and the match types are no longer necessary for advertisers. If the advertiser adopts the new bidding language, first, all the match type of his keywords will be extended to the advanced broad match, and then models that can retrieve related ads directly from the query can be used for intelligent targeting to break the limits of his selected keywords.  

The intelligent targeting service is an end-to-end ad retrieval framework and can achieve direct matching from query to related ads. The overall architecture of the intelligent targeting service is presented in Figure 4, and two types of ad retrieval models are used in this service which can provide more ad candidates by utilizing diverse data sources.   
\begin{figure}[h]
  \centering
  \includegraphics[width=\linewidth]{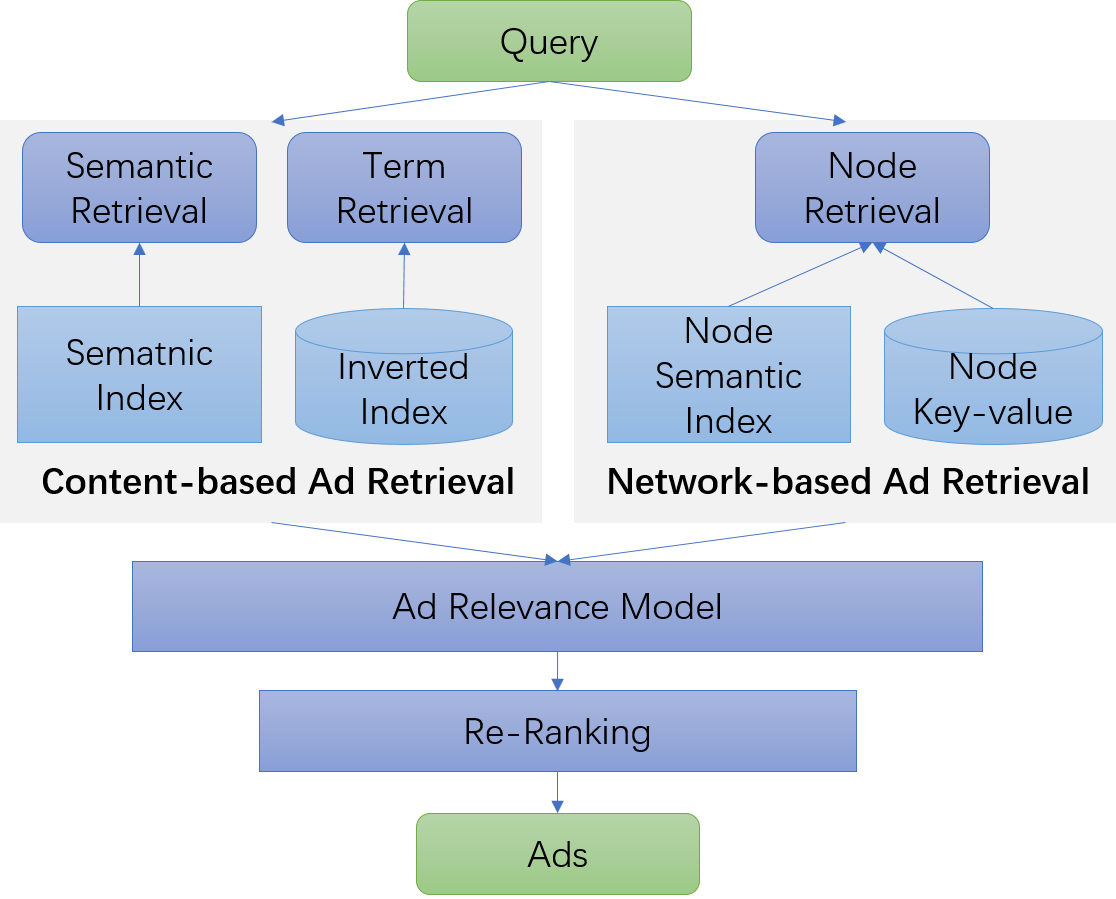}
  \caption{Architecture overview of the intelligent targeting service.}
  \Description{Architecture overview of the intelligent targeting service.}
\end{figure}  

\subsection{Network-based Ad Retrieval}

To retrieve related ads for search queries, a straightforward method is to utilize the multiple relationships between queries and ads. Based on the historical click data, a heterogeneous network can be constructed to encode structured information of multiple types of nodes and links. A snippet of this multi-relational network is presented in Figure 5. There are four types of nodes and seven types of relationships in this heterogeneous network. Based on the structure of this network, we can find the potential paths from queries to unconnected ads, such as the node Query4 and the node Ad4. 
\begin{figure}[h]
  \centering
  \includegraphics[width=7cm,height=3.5cm]{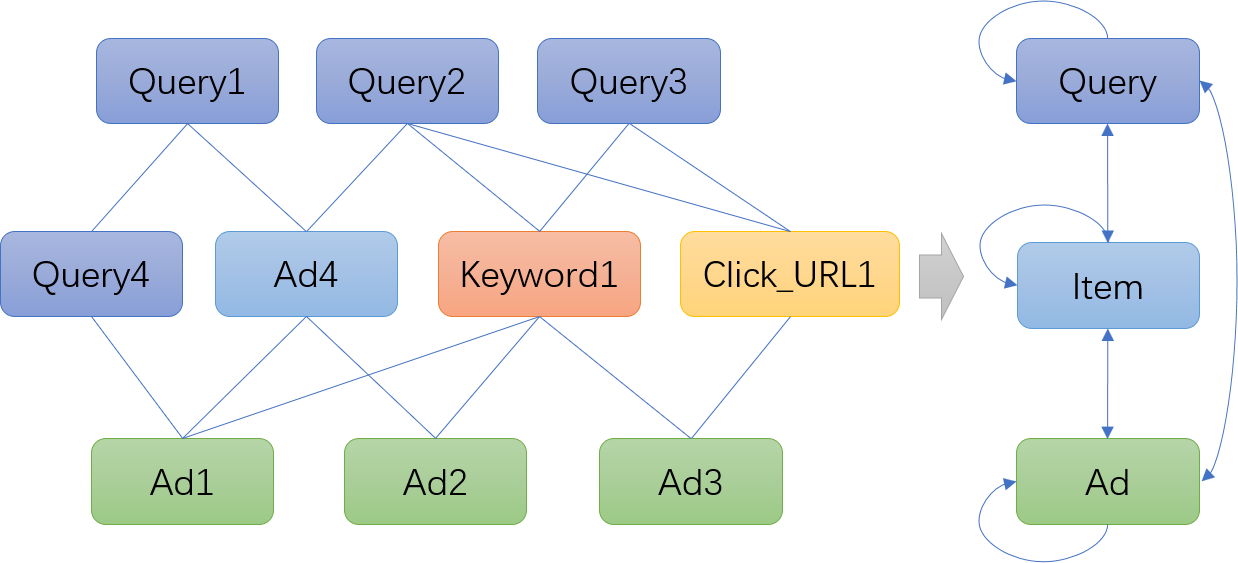}
  \caption{A snippet of the multi-relational networks.}
  \Description{A snippet of the multi-relational networks.}
\end{figure} 

A heterogeneous network is defined as a graph $G=(V,E,T)$ in which each node $v$ and each link $e$ are associated with their mapping functions$\phi(v): V \rightarrow T_V$ and $\varphi(e): E \rightarrow T_E$, respectively. $T_V$ and $T_E$ denote the sets of object and relation types, where $|T_V|+|T_E|>2$. 
A meta-path\cite{sun2011pathsim} $P$ is a path defined on the network schema $T = (T_V,T_E)$ which represents a compositional relations between two given types. Examples of meta-paths from query to ad defined in network schema Figure 5 include $query1 \rightarrow query4 \rightarrow ad1$, and $query2 \rightarrow keyword1 \rightarrow ad2 \rightarrow  ad4$. 

Firstly, to search top-K similar ad nodes for a given query node, we use the PathSim\cite{sun2011pathsim}, a meta-path-based similarity measure. Given a symmetric meta path $P$, PathSim between query $q$ and ad $a$ is:
\begin{equation}
  s(q, a) = \frac{2 * |{p_{q\to a}:p_{q\to a}\in P}|}{ |{p_{q\to q}:p_{q\to q}\in P}| + |{p_{a\to a}:p_{a\to a}\in P}|}             
\end{equation}
where $p_{q\to a}$ is a path instance between $q$ and $a$, $p_{q\to q}$ is that between $q$ and $q$, $p_{a\to a}$ is that between $a$ and $a$. $s(x, y)$ is defined in terms of two parts: their connectivity defined by the number of paths between them following P, and the balance of their visibility, where the visibility is defined as the number of path instances between themselves. We implement a distributed computation of the PathSim between queries and ads based on MapReduce framework, and the top-k results for each query can be used to build a node key-value index for online targeting service.

In the meantime, the network embedding model also can be used in node retrieval in the heterogeneous network. In our system, the  heterogeneous skip-gram model\cite{dong2017metapath2vec} is used to learn the latent vector representation for multiple types of nodes. The goal of heterogeneous skip-gram is to maximize the likelihood of preserving both the structures and semantics of a given heterogeneous network, which can be formulated as:
\begin{equation}
   \mathop{\arg\min}\limits_{\theta} \sum_{v \in V} \sum_{t \in T_V} \sum_{c_t \in N_{t}(v)} {log p(c_t|v;\theta)}  
\end{equation} 

where $N_t(v)$ denotes $v$’s neighborhood with the $t$th type of nodes and $p(c_t|v;\theta)$ is commonly defined as a softmax function, 
\begin{equation}
p(c_t|v;\theta) = \frac{e^{X_{c_t}.X_v}}{\sum_{u_t \in V_t} {e^{X_{c_t}.X_v}}}
\end{equation}
where $X_v$ is the $v$th row of $X$, representing the embedding vector for node $v$, and $V_t$ is the node set of type $t$ in the network.
We adopt the meta-path-based random walks\cite{dong2017metapath2vec} to generate paths of multiple types of nodes. Based on the heterogeneous negative sampling method, the optimization objective is:
\begin{equation}
O(X) = log \sigma(X_{c_t}.X_v) + \sum_{m=1}^{M}{E_{u_{t} ^ m} \sim P_t(u_t)[log \sigma(-X_{u_{t} ^ m} . X_v)]}
\end{equation}

Further than that, since the heterogeneous skip-gram model can't learn good representations for the long-tail and unseen nodes and can't sufficiently utilize the rich attributes of nodes, we also adopt the inductive learning framework called GraphSAGE\cite{hamilton2017inductive} and take a meta-path‑guided aggregation manner to solve the cold-start problem and to leverage node attribute information. 

Firstly, to utilize the information of node attributes, we extend the basic heterogeneous network to heterogeneous information network(HIN), which means that we take the text content as the attribute of different types of node. To capture the semantic information of each nodes and reduce the dimension of node embedding parameters, we use the term embedding to compose the node embedding vector with an average function $W_i = average(w_t1, w_t2,..., w_tn)$. The term embeddings are pre-trained with the unsupervised skip-gram model and fine-tuned with the representation learning of node embeddings. 

Secondly, for the HIN, the aggregation process should follow the guide of different meta paths. Given a meta-path $t={t_1, t_2, ...t_n}$, if $T(e1) = t_1$, so the first-order neighbors of $e1$ for the $t$ is $N_{t}^1(e1)=E_t(t_1 \to t_2)$, in the same way, we can get the $N_{t}^k(e1)$ follow the meta-path.
Following different meta-paths for queries and ads nodes, we can get the aggregation representation for query and ad from different paths in bottom-up manner, each layer can use shared weights, then we can use a fully-connected layer as the output $p_i = sigmoid(F(W_q, W_a))$. Then we can learn the embedding parameters with a log-loss function, and the positive instances are the query-ad pairs with clicks or conversions, the negative instances are the query-ad pairs without clicks. The model architecture of GraphSAGE for HIN is presented in Figure 6. 

\begin{figure}[h]
  \centering
  \includegraphics[width=\linewidth]{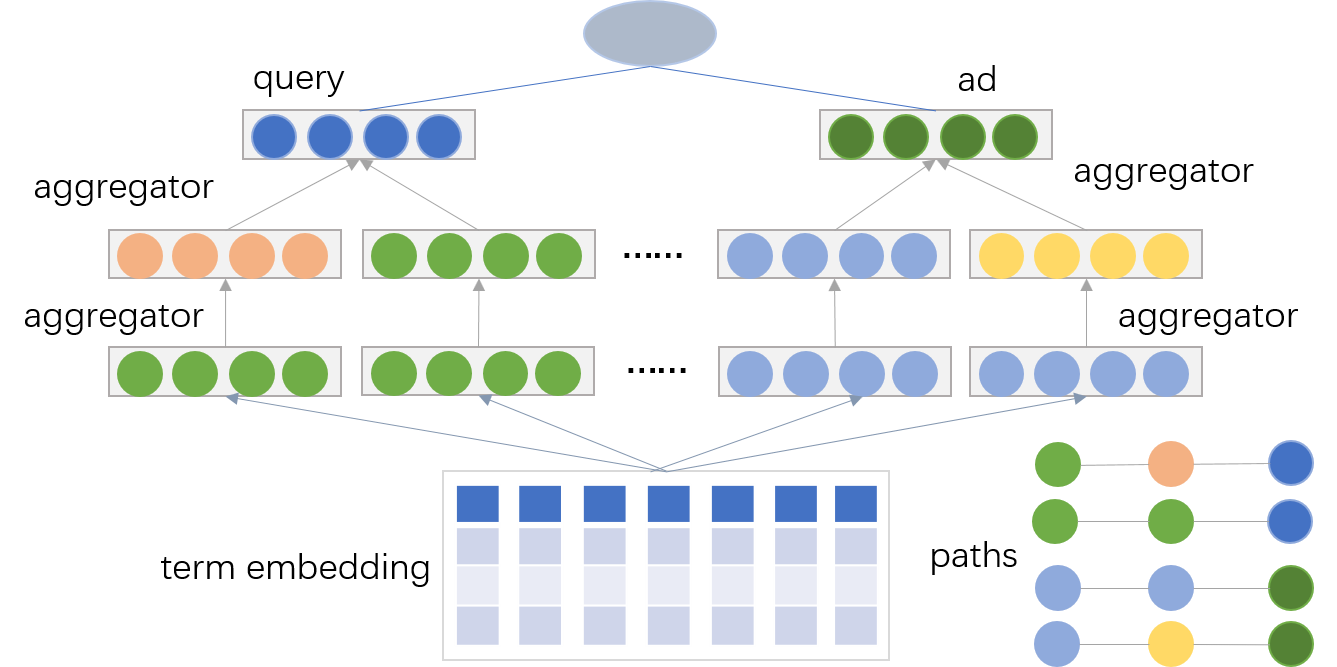}
  \caption{The architecture of GraphSAGE for HIN.}
  \Description{The architecture of GraphSAGE for HIN.}
\end{figure} 

The GraphSAGE model for HIN can also be used to generate node embeddings for previously unseen queries and ads nodes.

Based on the metapath2vec++ model and GraphSAGE model for HIN, we can learn the low-dimensional and latent embeddings for query nodes and ad nodes in the heterogeneous network. And the learned latent vectors for query and ad nodes can be used in building the node semantic index based on the Approximate Nearest Neighbor(ANN) retrieval platform.

Based on the structure of heterogeneous network built from historical click data, we can utilize the heterogeneous network mining method and heterogeneous network embedding model to find the potential relationship between queries and ads.

\subsection{Content-based Ad Retrieval}

Another straightforward retrieval model is content-based. As we can represent the query or the ad as a document, so the ad retrieval task can adopt a text-retrieval manner. In other words, we can retrieve related ads for target query just like the retrieving methods of organic search results.

At first, we should organize the queries and ads as the corresponding document as follows:
\begin{itemize}
\item {\verb|Query Docment|}: The query document consists of query text and its extensions such as clicked queries in the same session, clicked titles and snippets in search result and so on.
\item {\verb|Ad Docment|}: The ad document consists of ad text and its extensions such as contents extracted from landing page, clicked titles, clicked keywords, clicked creations and so on.
\end{itemize}

Based on query documents and ad documents, we developed term retrieval and semantic retrieval service. For the term retrieval, an inverted index is built based on the ad documents, and given a request query, we use the traditional term retrieval metrics to return top-k related ad documents as a traditional information retrieval(IR) task with relevance ranking model. 

For the semantic retrieval, deep learning models for the semantic matching problem can be used to learn the semantic representations of query docs and ad docs. In our system, we adopt the CDSSM\cite{shen2014learning,shen2014latent} model architecture which incorporates a convolutional-pooling structure over word sequences to learn low-dimensional, semantic vector representations for search queries and ad documents. By using the convolution-max pooling operation, local contextual information at the word n-gram level is modeled first. Then, salient local features in a word sequence are combined to form a global feature vector. Finally, the high-level semantic information of the word sequence is extracted to form a global vector representation. Some other advanced neural matching models are also can be adopted to learn the semantic representations. 

The semantic matching models are trained on click-through data by maximizing the conditional likelihood of clicked documents given a query. And the learned latent vectors for queries and ads can be used in building the semantic index based on the Approximate Nearest Neighbor(ANN) retrieval platform.

\section{Intelligent Creation Framework}
The sponsored search system provides many ad formats, and the advertisers have many materials for ad creation such as image, product links, phone number, app package and so on. Since the content and layout for ad creation is diverse and rich, the search space of ad creation is huge and it's very dizzy for advertisers to choose the ad formats or combinations of materials.

\begin{figure}[h]
  \centering
  \includegraphics[width=\linewidth]{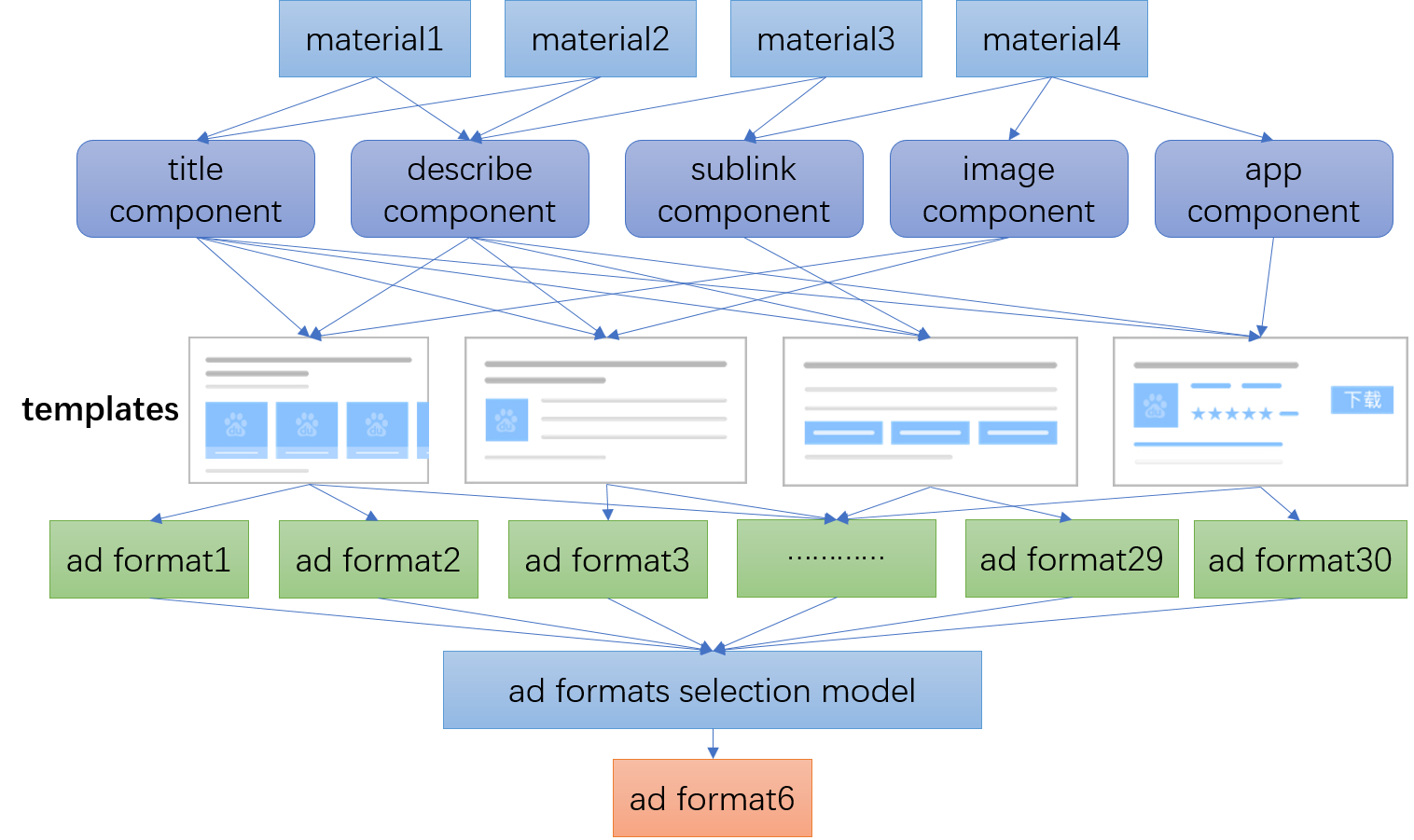}
  \caption{The intelligent creation framework.}
  \Description{The intelligent creation framework.}
\end{figure} 

To solve this problem, we design a componentized framework to design and generate ad creations intelligently. The overall architecture of the intelligent creation framework is presented in Figure 7. We treat the content and layout of each ad as a hierarchical 4-tuple(Material, Component, Template, Format).  

Based on this framework, the advertiser only needs to prepare the materials such as text description about their business, promotional products, images in different sizes, app package, phone number, and so on. The system will also extract and generate some materials from advertiser's landing page or other external resources. Based on these materials, the various ad components are created by the system to transform the materials into functional units such as a title component, a description component, a call button or a call link component, an image component, and so on. Then based on the available candidate sizes and layout, different templates are generated by searching the combination space automatically under the constraint of UE design rule. The ad template consists of many slots, each slot can be filled with a suitable ad component. So based on the generated ad templates, plenty of ad formats can be automatically generated by filling with ad components. 

Given the materials provided by advertisers, the componentized framework can generate different combinations and layouts automatically. The task of designing and generating ad templates can be formulated as the 2D rectangle packing problem which is NP-hard, we take a heuristic search algorithm\cite{jylanki2010thousand} and train an evaluation network to select valid templates with high CTR.

Finally, given plenty of generated ad formats, we design a selection model to predict the CTR and CVR for each ad format in different contexts. Given the target CPA, CTR and CVR, we then rank the ad formats based their expected cost-per-mille($eCPM = CPA * CVR * CTR$), and the ad format with the highest score will be selected to display.

\section{Experiments }

To evaluate and demonstrate the effectiveness of our proposed system, we conduct the online A/B test and long-term grouping experiment on live traffic of Baidu. As the AiAds system will take over the main campaign optimization task such as bidding, targeting, ad creation and the system relies on that the advertiser should use conversion tracking tools provided by Baidu\cite{Baiduconv19}, we need the authorization from advertisers. 

We collect 13350 advertisers from Baidu sponsored search platform, and set up the conversion tracking tools for them to get the conversions data. Then we choose 670 advertisers to use the AiAds system as the experimental group, and the others can be treated as the controlled group. Meanwhile, for the experimental group, we can also conduct the online A/B test to verify the performance of the AiAds system and each module. The statistics of the experimental group are presented in Table 1.

\begin{table}[!htbp]
  \caption{Information of the experiment group}
  \label{tab:freq}
  \begin{tabular}{ccl}
    \toprule
    Item&Numbers\\
    \midrule
    Advertisers & 670\\
    Daily impressions  & 101528334\\
    Daily clicks & 6528431\\
    Daily conversions & 378267\\
    Nodes & 119490684\\
    Edges & 359530239\\
  \bottomrule
\end{tabular}
\end{table}

For the experimental group, the advertisers adopt the target CPA bidding language and the automated bidding strategy, their keyword-level manual bidding is invalid. Their keyword match types are changed to broad match, and the intelligent targeting service provides additional ad retrieval results. They also can use the intelligent creation framework to optimize their ad creations by submitting multifarious materials, and the system generates more and more ad formats for them.

Firstly, the experimental results of the online A/B test are illustrated in table 2. We compared the AdAds system, the auto bidding strategy and the intelligent creation framework with the traditional baseline of manual optimization. The intelligent targeting service cannot be tested alone because it must be used in conjunction with the automated bidding strategy. The AiAds system can get more than 56\% improvement in conversions and more than 47\% improvement in the revenue of sponsored search system. Compared with the traditional advertising system, the AiAds has an overwhelming advantage due to its technical advancement. 

\begin{table}[!htbp]
  \caption{Experimental result of the online A/B test}
  \label{tab:freq}
  \begin{tabular}{ccccl}
    \toprule
    Metric&AiAds&Auto-Bidding&Intelligent Creation\\
    \midrule
    Revenue & 43.51\% & 24.49\% & 6.83\% \\
    Click & 26.88\% & 10.93\% & 6.77\% \\
    Conversion & 56.91\% & 38.35\% & 5.64\% \\
    CVR & 23.67\% & 24.72\% & -1.06\% \\
    CPA & -8.54\% & -10.02\% & 1.13\% \\
    Ad Quality & +0.89\% & +1.21\% & +0.03\% \\
  \bottomrule
\end{tabular}
\end{table}

Secondly, as the AiAds system has deployed and launched for a long time at Baidu, we can conduct the long-term grouping experiment. We trace and compare the metrics of advertisers in experiment group with the advertisers in the controlled group for a year. The comparison between the performance during the last week in 2018 with the performance during the last week in 2017 for these two groups is presented in table 3. From the result we can find that the revenue of sponsored search and budget of the advertisers in the experimental group have significantly increased.

\begin{table}[!htbp]
  \caption{Experimental result of the long-term grouping experiment}
  \label{tab:freq}
  \begin{tabular}{ccccl}
    \toprule
    Metric&Experiment Group&Controlled Group\\
    \midrule
    Revenue & 27.22\% & 10.85\%  \\
    Budget & 30.20\% & 8.83\%  \\
    Conversion & 22.67\% & 5.20\%  \\
    CVR & 18.24\% & -2.91\%  \\
    CPA & 3.71\% & 5.37\%  \\
  \bottomrule
\end{tabular}
\end{table}

Both the online A/B test and the long-term grouping experiment demonstrate the advantage and the effectiveness of the AiAds system. For Baidu, there are more and more advertisers are switching to using this new advertising system. Since the advertising platform is an auction market, the performance of AiAds will gradually abate as the increase of more advertisers who choose the AiAds, but it still has a distinct advantage compared with the baseline due to the lifting in CVR and finding new query-ad matching space.  

\section{Conclusions}

In this paper, we introduce the automated and intelligent advertising system deployed at Baidu. As the manual optimizations of keyword-level bidding, keyword selection and ad creation are highly inefficient and time-consuming, We design and develop automated bidding strategy, intelligent targeting model and intelligent creation framework to combine 3 optimization technologies to take the labor and guesswork out of targeting, bidding, and ad creation. 

The automated bidding strategy can replace the traditional keyword-level and manual bidding strategy, and given a simple and straight bidding language for advertisers. The intelligent targeting model can break the limits of keyword targeting and achieve straightforward ad retrieval from queries to ads, which will bring more commercial traffic to advertisers and sponsored search platform. The intelligent creation framework provides a simple way to have a global optimization for the ad creation and layout. 

By replacing manual tasks with automated and intelligent models, we think that the AiAds system opens a new door, and is bringing a revolution for sponsored search. For the whole digital advertising market, the intelligent technologies will reform or rebuild most parts of the market and will bring new opportunities and challenges. The underlying principle is letting the machines do what they do best. We hope that the lessons learned in building the AiAds system can also be helpful to the entire advertising market. 

Due to the limits of pages, we only give the guideline of this system but can't fully introduce the details for each model. Furthermore, the technologies behind the AiAds are still improving, we will separately introduce each part of AiAds in a more detailed way in the future. As the future works we will also optimize the ad retrieval model by utilizing more data sources and advanced models. Meanwhile, there are some open problems in designing the reasonable mechanism for ROI-constrained bidders.

%
% The acknowledgments section is defined using the "acks" environment (and NOT an unnumbered section). This ensures
% the proper identification of the section in the article metadata, and the consistent spelling of the heading.
\begin{acks}
We would sincerely like to thank the advertisers who participated in the early experiments of AiAds. We also thank the anonymous reviewers for their valuable comments and helpful suggestions.
\end{acks}

%
% The next two lines define the bibliography style to be used, and the bibliography file.
\bibliographystyle{ACM-Reference-Format}
\bibliography{kdd19_arxiv-bib}


\begin{thebibliography}{49}

%%% the Web address.

\ifx \showCODEN    \undefined \def \showCODEN     #1{\unskip}     \fi
\ifx \showDOI      \undefined \def \showDOI       #1{#1}\fi
\ifx \showISBNx    \undefined \def \showISBNx     #1{\unskip}     \fi
\ifx \showISBNxiii \undefined \def \showISBNxiii  #1{\unskip}     \fi
\ifx \showISSN     \undefined \def \showISSN      #1{\unskip}     \fi
\ifx \showLCCN     \undefined \def \showLCCN      #1{\unskip}     \fi
\ifx \shownote     \undefined \def \shownote      #1{#1}          \fi
\ifx \showarticletitle \undefined \def \showarticletitle #1{#1}   \fi
\ifx \showURL      \undefined \def \showURL       {\relax}        \fi
% invisible to TeX
\providecommand\bibfield[2]{#2}
\providecommand\bibinfo[2]{#2}
\providecommand\natexlab[1]{#1}
\providecommand\showeprint[2][]{arXiv:#2}

\bibitem[\protect\citeauthoryear{Agarwal, Ghosh, Wei, and You}{Agarwal
  et~al\mbox{.}}{2014}]%
        {agarwal2014budget}
\bibfield{author}{\bibinfo{person}{Deepak Agarwal}, \bibinfo{person}{Souvik
  Ghosh}, \bibinfo{person}{Kai Wei}, {and} \bibinfo{person}{Siyu You}.}
  \bibinfo{year}{2014}\natexlab{}.
\newblock \showarticletitle{Budget pacing for targeted online advertisements at
  linkedin}. In \bibinfo{booktitle}{\emph{Proceedings of the 20th ACM SIGKDD
  international conference on Knowledge discovery and data mining}}. ACM,
  \bibinfo{pages}{1613--1619}.
\newblock


\bibitem[\protect\citeauthoryear{Anastasakos, Hillard, Kshetramade, and
  Raghavan}{Anastasakos et~al\mbox{.}}{2009}]%
        {anastasakos2009collaborative}
\bibfield{author}{\bibinfo{person}{Tasos Anastasakos}, \bibinfo{person}{Dustin
  Hillard}, \bibinfo{person}{Sanjay Kshetramade}, {and} \bibinfo{person}{Hema
  Raghavan}.} \bibinfo{year}{2009}\natexlab{}.
\newblock \showarticletitle{A collaborative filtering approach to ad
  recommendation using the query-ad click graph}. In
  \bibinfo{booktitle}{\emph{Proceedings of the 18th ACM conference on
  Information and knowledge management}}. ACM, \bibinfo{pages}{1927--1930}.
\newblock


\bibitem[\protect\citeauthoryear{Bachrach, Ceppi, Kash, Key, and
  Kurokawa}{Bachrach et~al\mbox{.}}{2014}]%
        {bachrach2014optimising}
\bibfield{author}{\bibinfo{person}{Yoram Bachrach}, \bibinfo{person}{Sofia
  Ceppi}, \bibinfo{person}{Ian~A Kash}, \bibinfo{person}{Peter Key}, {and}
  \bibinfo{person}{David Kurokawa}.} \bibinfo{year}{2014}\natexlab{}.
\newblock \showarticletitle{Optimising trade-offs among stakeholders in ad
  auctions}. In \bibinfo{booktitle}{\emph{Proceedings of the fifteenth ACM
  conference on Economics and computation}}. ACM, \bibinfo{pages}{75--92}.
\newblock


\bibitem[\protect\citeauthoryear{Baidu}{Baidu}{2019a}]%
        {Baidu19}
\bibfield{author}{\bibinfo{person}{Baidu}.} \bibinfo{year}{2019}\natexlab{a}.
\newblock \bibinfo{title}{Baidu ad formats}.
\newblock
\newblock
\urldef\tempurl%
\url{http://yingxiao.baidu.com/new/home/product/product/id/50?ly=product_union_author_lists}
\showURL{%
Retrieved 2019 from \tempurl}


\bibitem[\protect\citeauthoryear{Baidu}{Baidu}{2019b}]%
        {Baiduconv19}
\bibfield{author}{\bibinfo{person}{Baidu}.} \bibinfo{year}{2019}\natexlab{b}.
\newblock \bibinfo{title}{Baidu conversion tracking}.
\newblock
\newblock
\urldef\tempurl%
\url{http://ocpc.baidu.com/developer/d/guide}
\showURL{%
Retrieved 2019 from \tempurl}


\bibitem[\protect\citeauthoryear{Bing}{Bing}{2019}]%
        {bingbid19}
\bibfield{author}{\bibinfo{person}{Bing}.} \bibinfo{year}{2019}\natexlab{}.
\newblock \bibinfo{title}{Bing automated bid strategies}.
\newblock
\newblock
\urldef\tempurl%
\url{https://help.bingads.microsoft.com/apex/index/3/en/56786}
\showURL{%
Retrieved 2019 from \tempurl}


\bibitem[\protect\citeauthoryear{Broder, Ciccolo, Gabrilovich, Josifovski,
  Metzler, Riedel, and Yuan}{Broder et~al\mbox{.}}{2009}]%
        {broder2009online}
\bibfield{author}{\bibinfo{person}{Andrei Broder}, \bibinfo{person}{Peter
  Ciccolo}, \bibinfo{person}{Evgeniy Gabrilovich}, \bibinfo{person}{Vanja
  Josifovski}, \bibinfo{person}{Donald Metzler}, \bibinfo{person}{Lance
  Riedel}, {and} \bibinfo{person}{Jeffrey Yuan}.}
  \bibinfo{year}{2009}\natexlab{}.
\newblock \showarticletitle{Online expansion of rare queries for sponsored
  search}. In \bibinfo{booktitle}{\emph{Proceedings of the 18th international
  conference on World wide web}}. ACM, \bibinfo{pages}{511--520}.
\newblock


\bibitem[\protect\citeauthoryear{Broder, Gabrilovich, Josifovski, Mavromatis,
  and Smola}{Broder et~al\mbox{.}}{2011}]%
        {broder2011bid}
\bibfield{author}{\bibinfo{person}{Andrei Broder}, \bibinfo{person}{Evgeniy
  Gabrilovich}, \bibinfo{person}{Vanja Josifovski}, \bibinfo{person}{George
  Mavromatis}, {and} \bibinfo{person}{Alex Smola}.}
  \bibinfo{year}{2011}\natexlab{}.
\newblock \showarticletitle{Bid generation for advanced match in sponsored
  search}. In \bibinfo{booktitle}{\emph{Proceedings of the fourth ACM
  international conference on Web search and data mining}}. ACM,
  \bibinfo{pages}{515--524}.
\newblock


\bibitem[\protect\citeauthoryear{Broder, Ciccolo, Fontoura, Gabrilovich,
  Josifovski, and Riedel}{Broder et~al\mbox{.}}{2008}]%
        {broder2008search}
\bibfield{author}{\bibinfo{person}{Andrei~Z Broder}, \bibinfo{person}{Peter
  Ciccolo}, \bibinfo{person}{Marcus Fontoura}, \bibinfo{person}{Evgeniy
  Gabrilovich}, \bibinfo{person}{Vanja Josifovski}, {and}
  \bibinfo{person}{Lance Riedel}.} \bibinfo{year}{2008}\natexlab{}.
\newblock \showarticletitle{Search advertising using web relevance feedback}.
  In \bibinfo{booktitle}{\emph{Proceedings of the 17th ACM conference on
  Information and knowledge management}}. ACM, \bibinfo{pages}{1013--1022}.
\newblock


\bibitem[\protect\citeauthoryear{Cary, Das, Edelman, Giotis, Heimerl, Karlin,
  Mathieu, and Schwarz}{Cary et~al\mbox{.}}{2007}]%
        {cary2007greedy}
\bibfield{author}{\bibinfo{person}{Matthew Cary}, \bibinfo{person}{Aparna Das},
  \bibinfo{person}{Ben Edelman}, \bibinfo{person}{Ioannis Giotis},
  \bibinfo{person}{Kurtis Heimerl}, \bibinfo{person}{Anna~R Karlin},
  \bibinfo{person}{Claire Mathieu}, {and} \bibinfo{person}{Michael Schwarz}.}
  \bibinfo{year}{2007}\natexlab{}.
\newblock \showarticletitle{Greedy bidding strategies for keyword auctions}. In
  \bibinfo{booktitle}{\emph{Proceedings of the 8th ACM conference on Electronic
  commerce}}. ACM, \bibinfo{pages}{262--271}.
\newblock


\bibitem[\protect\citeauthoryear{Cavallo, Krishnamurthy, Sviridenko, and
  Wilkens}{Cavallo et~al\mbox{.}}{2017}]%
        {cavallo2017sponsored}
\bibfield{author}{\bibinfo{person}{Ruggiero Cavallo},
  \bibinfo{person}{Prabhakar Krishnamurthy}, \bibinfo{person}{Maxim
  Sviridenko}, {and} \bibinfo{person}{Christopher~A Wilkens}.}
  \bibinfo{year}{2017}\natexlab{}.
\newblock \showarticletitle{Sponsored search auctions with rich ads}. In
  \bibinfo{booktitle}{\emph{Proceedings of the 26th International Conference on
  World Wide Web}}. International World Wide Web Conferences Steering
  Committee, \bibinfo{pages}{43--51}.
\newblock


\bibitem[\protect\citeauthoryear{Choi, Fontoura, Gabrilovich, Josifovski,
  Mediano, and Pang}{Choi et~al\mbox{.}}{2010}]%
        {choi2010using}
\bibfield{author}{\bibinfo{person}{Yejin Choi}, \bibinfo{person}{Marcus
  Fontoura}, \bibinfo{person}{Evgeniy Gabrilovich}, \bibinfo{person}{Vanja
  Josifovski}, \bibinfo{person}{Mauricio Mediano}, {and} \bibinfo{person}{Bo
  Pang}.} \bibinfo{year}{2010}\natexlab{}.
\newblock \showarticletitle{Using landing pages for sponsored search ad
  selection}. In \bibinfo{booktitle}{\emph{Proceedings of the 19th
  international conference on World wide web}}. ACM, \bibinfo{pages}{251--260}.
\newblock


\bibitem[\protect\citeauthoryear{Dong, Chawla, and Swami}{Dong
  et~al\mbox{.}}{2017}]%
        {dong2017metapath2vec}
\bibfield{author}{\bibinfo{person}{Yuxiao Dong}, \bibinfo{person}{Nitesh~V
  Chawla}, {and} \bibinfo{person}{Ananthram Swami}.}
  \bibinfo{year}{2017}\natexlab{}.
\newblock \showarticletitle{metapath2vec: Scalable representation learning for
  heterogeneous networks}. In \bibinfo{booktitle}{\emph{Proceedings of the 23rd
  ACM SIGKDD International Conference on Knowledge Discovery and Data Mining}}.
  ACM, \bibinfo{pages}{135--144}.
\newblock


\bibitem[\protect\citeauthoryear{Even~Dar, Mirrokni, Muthukrishnan, Mansour,
  and Nadav}{Even~Dar et~al\mbox{.}}{2009}]%
        {even2009bid}
\bibfield{author}{\bibinfo{person}{Eyal Even~Dar}, \bibinfo{person}{Vahab~S
  Mirrokni}, \bibinfo{person}{S Muthukrishnan}, \bibinfo{person}{Yishay
  Mansour}, {and} \bibinfo{person}{Uri Nadav}.}
  \bibinfo{year}{2009}\natexlab{}.
\newblock \showarticletitle{Bid optimization for broad match ad auctions}. In
  \bibinfo{booktitle}{\emph{Proceedings of the 18th international conference on
  World wide web}}. ACM, \bibinfo{pages}{231--240}.
\newblock


\bibitem[\protect\citeauthoryear{Feldman and Muthukrishnan}{Feldman and
  Muthukrishnan}{2008}]%
        {feldman2008algorithmic}
\bibfield{author}{\bibinfo{person}{Jon Feldman} {and} \bibinfo{person}{S
  Muthukrishnan}.} \bibinfo{year}{2008}\natexlab{}.
\newblock \showarticletitle{Algorithmic methods for sponsored search
  advertising}.
\newblock In \bibinfo{booktitle}{\emph{Performance Modeling and Engineering}}.
  \bibinfo{publisher}{Springer}, \bibinfo{pages}{91--122}.
\newblock


\bibitem[\protect\citeauthoryear{Golrezaei, Lobel, and Paes~Leme}{Golrezaei
  et~al\mbox{.}}{2018}]%
        {golrezaei2018auction}
\bibfield{author}{\bibinfo{person}{Negin Golrezaei}, \bibinfo{person}{Ilan
  Lobel}, {and} \bibinfo{person}{Renato Paes~Leme}.}
  \bibinfo{year}{2018}\natexlab{}.
\newblock \showarticletitle{Auction Design for ROI-Constrained Buyers}.
\newblock  (\bibinfo{year}{2018}).
\newblock


\bibitem[\protect\citeauthoryear{Google}{Google}{2019a}]%
        {Google19}
\bibfield{author}{\bibinfo{person}{Google}.} \bibinfo{year}{2019}\natexlab{a}.
\newblock \bibinfo{title}{Google ads extensions}.
\newblock
\newblock
\urldef\tempurl%
\url{https://support.google.com/google-ads/answer/2375499}
\showURL{%
Retrieved 2019 from \tempurl}


\bibitem[\protect\citeauthoryear{Google}{Google}{2019b}]%
        {googlebid19}
\bibfield{author}{\bibinfo{person}{Google}.} \bibinfo{year}{2019}\natexlab{b}.
\newblock \bibinfo{title}{Google automated bid strategies}.
\newblock
\newblock
\urldef\tempurl%
\url{https://support.google.com/google-ads/answer/2979071?hl=en}
\showURL{%
Retrieved 2019 from \tempurl}


\bibitem[\protect\citeauthoryear{Grbovic, Djuric, Radosavljevic, Silvestri,
  Baeza-Yates, Feng, Ordentlich, Yang, and Owens}{Grbovic
  et~al\mbox{.}}{2016}]%
        {grbovic2016scalable}
\bibfield{author}{\bibinfo{person}{Mihajlo Grbovic}, \bibinfo{person}{Nemanja
  Djuric}, \bibinfo{person}{Vladan Radosavljevic}, \bibinfo{person}{Fabrizio
  Silvestri}, \bibinfo{person}{Ricardo Baeza-Yates}, \bibinfo{person}{Andrew
  Feng}, \bibinfo{person}{Erik Ordentlich}, \bibinfo{person}{Lee Yang}, {and}
  \bibinfo{person}{Gavin Owens}.} \bibinfo{year}{2016}\natexlab{}.
\newblock \showarticletitle{Scalable semantic matching of queries to ads in
  sponsored search advertising}. In \bibinfo{booktitle}{\emph{Proceedings of
  the 39th International ACM SIGIR conference on Research and Development in
  Information Retrieval}}. ACM, \bibinfo{pages}{375--384}.
\newblock


\bibitem[\protect\citeauthoryear{Gupta, Bilenko, and Richardson}{Gupta
  et~al\mbox{.}}{2009}]%
        {gupta2009catching}
\bibfield{author}{\bibinfo{person}{Sonal Gupta}, \bibinfo{person}{Mikhail
  Bilenko}, {and} \bibinfo{person}{Matthew Richardson}.}
  \bibinfo{year}{2009}\natexlab{}.
\newblock \showarticletitle{Catching the drift: learning broad matches from
  clickthrough data}. In \bibinfo{booktitle}{\emph{Proceedings of the 15th ACM
  SIGKDD international conference on Knowledge discovery and data mining}}.
  ACM, \bibinfo{pages}{1165--1174}.
\newblock


\bibitem[\protect\citeauthoryear{Hamilton, Ying, and Leskovec}{Hamilton
  et~al\mbox{.}}{2017}]%
        {hamilton2017inductive}
\bibfield{author}{\bibinfo{person}{Will Hamilton}, \bibinfo{person}{Zhitao
  Ying}, {and} \bibinfo{person}{Jure Leskovec}.}
  \bibinfo{year}{2017}\natexlab{}.
\newblock \showarticletitle{Inductive representation learning on large graphs}.
  In \bibinfo{booktitle}{\emph{Advances in Neural Information Processing
  Systems}}. \bibinfo{pages}{1024--1034}.
\newblock


\bibitem[\protect\citeauthoryear{Hartline, Immorlica, Khani, Lucier, and
  Niazadeh}{Hartline et~al\mbox{.}}{2018}]%
        {hartline2018fast}
\bibfield{author}{\bibinfo{person}{Jason Hartline}, \bibinfo{person}{Nicole
  Immorlica}, \bibinfo{person}{Mohammad~Reza Khani}, \bibinfo{person}{Brendan
  Lucier}, {and} \bibinfo{person}{Rad Niazadeh}.}
  \bibinfo{year}{2018}\natexlab{}.
\newblock \showarticletitle{Fast Core Pricing for Rich Advertising Auctions}.
  In \bibinfo{booktitle}{\emph{Proceedings of the 2018 ACM Conference on
  Economics and Computation}}. ACM, \bibinfo{pages}{111--112}.
\newblock


\bibitem[\protect\citeauthoryear{Heymann}{Heymann}{2018}]%
        {heymann2018roi}
\bibfield{author}{\bibinfo{person}{Benamin Heymann}.}
  \bibinfo{year}{2018}\natexlab{}.
\newblock \showarticletitle{ROI constrained Auctions}.
\newblock \bibinfo{journal}{\emph{arXiv preprint arXiv:1809.08837}}
  (\bibinfo{year}{2018}).
\newblock


\bibitem[\protect\citeauthoryear{Jyl{\"a}nki}{Jyl{\"a}nki}{2010}]%
        {jylanki2010thousand}
\bibfield{author}{\bibinfo{person}{Jukka Jyl{\"a}nki}.}
  \bibinfo{year}{2010}\natexlab{}.
\newblock \showarticletitle{A thousand ways to pack the bin-a practical
  approach to two-dimensional rectangle bin packing}.
\newblock \bibinfo{journal}{\emph{retrived from http://clb. demon.
  fi/files/RectangleBinPack. pdf}} (\bibinfo{year}{2010}).
\newblock


\bibitem[\protect\citeauthoryear{Ke, Meng, Finley, Wang, Chen, Ma, Ye, and
  Liu}{Ke et~al\mbox{.}}{2017}]%
        {ke2017lightgbm}
\bibfield{author}{\bibinfo{person}{Guolin Ke}, \bibinfo{person}{Qi Meng},
  \bibinfo{person}{Thomas Finley}, \bibinfo{person}{Taifeng Wang},
  \bibinfo{person}{Wei Chen}, \bibinfo{person}{Weidong Ma},
  \bibinfo{person}{Qiwei Ye}, {and} \bibinfo{person}{Tie-Yan Liu}.}
  \bibinfo{year}{2017}\natexlab{}.
\newblock \showarticletitle{Lightgbm: A highly efficient gradient boosting
  decision tree}. In \bibinfo{booktitle}{\emph{Advances in Neural Information
  Processing Systems}}. \bibinfo{pages}{3146--3154}.
\newblock


\bibitem[\protect\citeauthoryear{Lee, Jalali, and Dasdan}{Lee
  et~al\mbox{.}}{2013}]%
        {lee2013real}
\bibfield{author}{\bibinfo{person}{Kuang-Chih Lee}, \bibinfo{person}{Ali
  Jalali}, {and} \bibinfo{person}{Ali Dasdan}.}
  \bibinfo{year}{2013}\natexlab{}.
\newblock \showarticletitle{Real time bid optimization with smooth budget
  delivery in online advertising}. In \bibinfo{booktitle}{\emph{Proceedings of
  the Seventh International Workshop on Data Mining for Online Advertising}}.
  ACM, \bibinfo{pages}{1}.
\newblock


\bibitem[\protect\citeauthoryear{Mirrokni, Leme, Tang, and Zuo}{Mirrokni
  et~al\mbox{.}}{2016}]%
        {mirrokni2016dynamic}
\bibfield{author}{\bibinfo{person}{Vahab Mirrokni},
  \bibinfo{person}{Renato~Paes Leme}, \bibinfo{person}{Pingzhong Tang}, {and}
  \bibinfo{person}{Song Zuo}.} \bibinfo{year}{2016}\natexlab{}.
\newblock \showarticletitle{Dynamic auctions with bank accounts}. In
  \bibinfo{booktitle}{\emph{Proceedings of the Twenty-Fifth International Joint
  Conference on Artificial Intelligence}}. AAAI Press,
  \bibinfo{pages}{387--393}.
\newblock


\bibitem[\protect\citeauthoryear{Mirrokni, Paes~Leme, Tang, and Zuo}{Mirrokni
  et~al\mbox{.}}{2017}]%
        {mirrokni2017non}
\bibfield{author}{\bibinfo{person}{Vahab Mirrokni}, \bibinfo{person}{Renato
  Paes~Leme}, \bibinfo{person}{Pingzhong Tang}, {and} \bibinfo{person}{Song
  Zuo}.} \bibinfo{year}{2017}\natexlab{}.
\newblock \showarticletitle{Non-clairvoyant dynamic mechanism design}.
\newblock  (\bibinfo{year}{2017}).
\newblock


\bibitem[\protect\citeauthoryear{Mnih, Kavukcuoglu, Silver, Graves, Antonoglou,
  Wierstra, and Riedmiller}{Mnih et~al\mbox{.}}{2013}]%
        {mnih2013playing}
\bibfield{author}{\bibinfo{person}{Volodymyr Mnih}, \bibinfo{person}{Koray
  Kavukcuoglu}, \bibinfo{person}{David Silver}, \bibinfo{person}{Alex Graves},
  \bibinfo{person}{Ioannis Antonoglou}, \bibinfo{person}{Daan Wierstra}, {and}
  \bibinfo{person}{Martin Riedmiller}.} \bibinfo{year}{2013}\natexlab{}.
\newblock \showarticletitle{Playing atari with deep reinforcement learning}.
\newblock \bibinfo{journal}{\emph{arXiv preprint arXiv:1312.5602}}
  (\bibinfo{year}{2013}).
\newblock


\bibitem[\protect\citeauthoryear{Mnih, Kavukcuoglu, Silver, Rusu, Veness,
  Bellemare, Graves, Riedmiller, Fidjeland, Ostrovski, et~al\mbox{.}}{Mnih
  et~al\mbox{.}}{2015}]%
        {mnih2015human}
\bibfield{author}{\bibinfo{person}{Volodymyr Mnih}, \bibinfo{person}{Koray
  Kavukcuoglu}, \bibinfo{person}{David Silver}, \bibinfo{person}{Andrei~A
  Rusu}, \bibinfo{person}{Joel Veness}, \bibinfo{person}{Marc~G Bellemare},
  \bibinfo{person}{Alex Graves}, \bibinfo{person}{Martin Riedmiller},
  \bibinfo{person}{Andreas~K Fidjeland}, \bibinfo{person}{Georg Ostrovski},
  {et~al\mbox{.}}} \bibinfo{year}{2015}\natexlab{}.
\newblock \showarticletitle{Human-level control through deep reinforcement
  learning}.
\newblock \bibinfo{journal}{\emph{Nature}} \bibinfo{volume}{518},
  \bibinfo{number}{7540} (\bibinfo{year}{2015}), \bibinfo{pages}{529}.
\newblock


\bibitem[\protect\citeauthoryear{Pandey, Punera, Fontoura, and
  Josifovski}{Pandey et~al\mbox{.}}{2010}]%
        {pandey2010estimating}
\bibfield{author}{\bibinfo{person}{Sandeep Pandey}, \bibinfo{person}{Kunal
  Punera}, \bibinfo{person}{Marcus Fontoura}, {and} \bibinfo{person}{Vanja
  Josifovski}.} \bibinfo{year}{2010}\natexlab{}.
\newblock \showarticletitle{Estimating advertisability of tail queries for
  sponsored search}. In \bibinfo{booktitle}{\emph{Proceedings of the 33rd
  international ACM SIGIR conference on Research and development in information
  retrieval}}. ACM, \bibinfo{pages}{563--570}.
\newblock


\bibitem[\protect\citeauthoryear{Qin, Chen, and Liu}{Qin et~al\mbox{.}}{2015}]%
        {qin2015sponsored}
\bibfield{author}{\bibinfo{person}{Tao Qin}, \bibinfo{person}{Wei Chen}, {and}
  \bibinfo{person}{Tie-Yan Liu}.} \bibinfo{year}{2015}\natexlab{}.
\newblock \showarticletitle{Sponsored search auctions: Recent advances and
  future directions}.
\newblock \bibinfo{journal}{\emph{ACM Transactions on Intelligent Systems and
  Technology (TIST)}} \bibinfo{volume}{5}, \bibinfo{number}{4}
  (\bibinfo{year}{2015}), \bibinfo{pages}{60}.
\newblock


\bibitem[\protect\citeauthoryear{Ravi, Broder, Gabrilovich, Josifovski, Pandey,
  and Pang}{Ravi et~al\mbox{.}}{2010}]%
        {ravi2010automatic}
\bibfield{author}{\bibinfo{person}{Sujith Ravi}, \bibinfo{person}{Andrei
  Broder}, \bibinfo{person}{Evgeniy Gabrilovich}, \bibinfo{person}{Vanja
  Josifovski}, \bibinfo{person}{Sandeep Pandey}, {and} \bibinfo{person}{Bo
  Pang}.} \bibinfo{year}{2010}\natexlab{}.
\newblock \showarticletitle{Automatic generation of bid phrases for online
  advertising}. In \bibinfo{booktitle}{\emph{Proceedings of the third ACM
  international conference on Web search and data mining}}. ACM,
  \bibinfo{pages}{341--350}.
\newblock


\bibitem[\protect\citeauthoryear{Rey and Kannan}{Rey and Kannan}{2010}]%
        {rey2010conversion}
\bibfield{author}{\bibinfo{person}{Benjamin Rey} {and} \bibinfo{person}{Ashvin
  Kannan}.} \bibinfo{year}{2010}\natexlab{}.
\newblock \showarticletitle{Conversion rate based bid adjustment for sponsored
  search}. In \bibinfo{booktitle}{\emph{Proceedings of the 19th international
  conference on World wide web}}. ACM, \bibinfo{pages}{1173--1174}.
\newblock


\bibitem[\protect\citeauthoryear{Shen, He, Gao, Deng, and Mesnil}{Shen
  et~al\mbox{.}}{2014a}]%
        {shen2014latent}
\bibfield{author}{\bibinfo{person}{Yelong Shen}, \bibinfo{person}{Xiaodong He},
  \bibinfo{person}{Jianfeng Gao}, \bibinfo{person}{Li Deng}, {and}
  \bibinfo{person}{Gr{\'e}goire Mesnil}.} \bibinfo{year}{2014}\natexlab{a}.
\newblock \showarticletitle{A latent semantic model with convolutional-pooling
  structure for information retrieval}. In
  \bibinfo{booktitle}{\emph{Proceedings of the 23rd ACM International
  Conference on Conference on Information and Knowledge Management}}. ACM,
  \bibinfo{pages}{101--110}.
\newblock


\bibitem[\protect\citeauthoryear{Shen, He, Gao, Deng, and Mesnil}{Shen
  et~al\mbox{.}}{2014b}]%
        {shen2014learning}
\bibfield{author}{\bibinfo{person}{Yelong Shen}, \bibinfo{person}{Xiaodong He},
  \bibinfo{person}{Jianfeng Gao}, \bibinfo{person}{Li Deng}, {and}
  \bibinfo{person}{Gr{\'e}goire Mesnil}.} \bibinfo{year}{2014}\natexlab{b}.
\newblock \showarticletitle{Learning semantic representations using
  convolutional neural networks for web search}. In
  \bibinfo{booktitle}{\emph{Proceedings of the 23rd International Conference on
  World Wide Web}}. ACM, \bibinfo{pages}{373--374}.
\newblock


\bibitem[\protect\citeauthoryear{Song, Kim, Park, et~al\mbox{.}}{Song
  et~al\mbox{.}}{2017}]%
        {song2017translation}
\bibfield{author}{\bibinfo{person}{Hyun-Je Song}, \bibinfo{person}{A Kim},
  \bibinfo{person}{Seong-Bae Park}, {et~al\mbox{.}}}
  \bibinfo{year}{2017}\natexlab{}.
\newblock \showarticletitle{Translation of Natural Language Query Into Keyword
  Query Using a RNN Encoder-Decoder}. In \bibinfo{booktitle}{\emph{Proceedings
  of the 40th International ACM SIGIR Conference on Research and Development in
  Information Retrieval}}. ACM, \bibinfo{pages}{965--968}.
\newblock


\bibitem[\protect\citeauthoryear{Statista}{Statista}{2019}]%
        {Statista19}
\bibfield{author}{\bibinfo{person}{Statista}.} \bibinfo{year}{2019}\natexlab{}.
\newblock \bibinfo{title}{Search Advertising - worldwide | Statista Market
  Forecast}.
\newblock
\newblock
\urldef\tempurl%
\url{https://www.statista.com/outlook/219/100/search-advertising/worldwide}
\showURL{%
Retrieved 2019 from \tempurl}


\bibitem[\protect\citeauthoryear{Sun, Han, Yan, Yu, and Wu}{Sun
  et~al\mbox{.}}{2011}]%
        {sun2011pathsim}
\bibfield{author}{\bibinfo{person}{Yizhou Sun}, \bibinfo{person}{Jiawei Han},
  \bibinfo{person}{Xifeng Yan}, \bibinfo{person}{Philip~S Yu}, {and}
  \bibinfo{person}{Tianyi Wu}.} \bibinfo{year}{2011}\natexlab{}.
\newblock \showarticletitle{Pathsim: Meta path-based top-k similarity search in
  heterogeneous information networks}.
\newblock \bibinfo{journal}{\emph{Proceedings of the VLDB Endowment}}
  \bibinfo{volume}{4}, \bibinfo{number}{11} (\bibinfo{year}{2011}),
  \bibinfo{pages}{992--1003}.
\newblock


\bibitem[\protect\citeauthoryear{Wilkens, Cavallo, and Niazadeh}{Wilkens
  et~al\mbox{.}}{2016}]%
        {wilkens2016mechanism}
\bibfield{author}{\bibinfo{person}{Christopher~A Wilkens},
  \bibinfo{person}{Ruggiero Cavallo}, {and} \bibinfo{person}{Rad Niazadeh}.}
  \bibinfo{year}{2016}\natexlab{}.
\newblock \showarticletitle{Mechanism design for value maximizers}.
\newblock \bibinfo{journal}{\emph{arXiv preprint arXiv:1607.04362}}
  (\bibinfo{year}{2016}).
\newblock


\bibitem[\protect\citeauthoryear{Xu, Lee, Li, Qi, and Lu}{Xu
  et~al\mbox{.}}{2015}]%
        {xu2015smart}
\bibfield{author}{\bibinfo{person}{Jian Xu}, \bibinfo{person}{Kuang-chih Lee},
  \bibinfo{person}{Wentong Li}, \bibinfo{person}{Hang Qi}, {and}
  \bibinfo{person}{Quan Lu}.} \bibinfo{year}{2015}\natexlab{}.
\newblock \showarticletitle{Smart pacing for effective online ad campaign
  optimization}. In \bibinfo{booktitle}{\emph{Proceedings of the 21th ACM
  SIGKDD International Conference on Knowledge Discovery and Data Mining}}.
  ACM, \bibinfo{pages}{2217--2226}.
\newblock


\bibitem[\protect\citeauthoryear{Yan, Lin, Wu, Xiao, Zheng, Wu, and Liu}{Yan
  et~al\mbox{.}}{2017}]%
        {yan2017beyond}
\bibfield{author}{\bibinfo{person}{Su Yan}, \bibinfo{person}{Wei Lin},
  \bibinfo{person}{Tianshu Wu}, \bibinfo{person}{Daorui Xiao},
  \bibinfo{person}{Xu Zheng}, \bibinfo{person}{Bo Wu}, {and}
  \bibinfo{person}{Kaipeng Liu}.} \bibinfo{year}{2017}\natexlab{}.
\newblock \showarticletitle{Beyond Keywords and Relevance: A Personalized Ad
  Retrieval Framework in E-Commerce Sponsored Search}.
\newblock \bibinfo{journal}{\emph{arXiv preprint arXiv:1712.10110}}
  (\bibinfo{year}{2017}).
\newblock


\bibitem[\protect\citeauthoryear{Zadrozny and Elkan}{Zadrozny and
  Elkan}{2002}]%
        {zadrozny2002transforming}
\bibfield{author}{\bibinfo{person}{Bianca Zadrozny} {and}
  \bibinfo{person}{Charles Elkan}.} \bibinfo{year}{2002}\natexlab{}.
\newblock \showarticletitle{Transforming classifier scores into accurate
  multiclass probability estimates}. In \bibinfo{booktitle}{\emph{Proceedings
  of the eighth ACM SIGKDD international conference on Knowledge discovery and
  data mining}}. ACM, \bibinfo{pages}{694--699}.
\newblock


\bibitem[\protect\citeauthoryear{Zhang, Wang, Xue, and Zha}{Zhang
  et~al\mbox{.}}{2012a}]%
        {zhang2012advertising}
\bibfield{author}{\bibinfo{person}{Weinan Zhang}, \bibinfo{person}{Dingquan
  Wang}, \bibinfo{person}{Gui-Rong Xue}, {and} \bibinfo{person}{Hongyuan Zha}.}
  \bibinfo{year}{2012}\natexlab{a}.
\newblock \showarticletitle{Advertising keywords recommendation for short-text
  web pages using Wikipedia}.
\newblock \bibinfo{journal}{\emph{ACM Transactions on Intelligent Systems and
  Technology (TIST)}} \bibinfo{volume}{3}, \bibinfo{number}{2}
  (\bibinfo{year}{2012}), \bibinfo{pages}{36}.
\newblock


\bibitem[\protect\citeauthoryear{Zhang, Zhang, Gao, Yu, Yuan, and Liu}{Zhang
  et~al\mbox{.}}{2012b}]%
        {zhang2012joint}
\bibfield{author}{\bibinfo{person}{Weinan Zhang}, \bibinfo{person}{Ying Zhang},
  \bibinfo{person}{Bin Gao}, \bibinfo{person}{Yong Yu},
  \bibinfo{person}{Xiaojie Yuan}, {and} \bibinfo{person}{Tie-Yan Liu}.}
  \bibinfo{year}{2012}\natexlab{b}.
\newblock \showarticletitle{Joint optimization of bid and budget allocation in
  sponsored search}. In \bibinfo{booktitle}{\emph{Proceedings of the 18th ACM
  SIGKDD international conference on Knowledge discovery and data mining}}.
  ACM, \bibinfo{pages}{1177--1185}.
\newblock


\bibitem[\protect\citeauthoryear{Zhang, He, Rey, and Jones}{Zhang
  et~al\mbox{.}}{2007}]%
        {zhang2007query}
\bibfield{author}{\bibinfo{person}{Wei~Vivian Zhang}, \bibinfo{person}{Xiaofei
  He}, \bibinfo{person}{Benjamin Rey}, {and} \bibinfo{person}{Rosie Jones}.}
  \bibinfo{year}{2007}\natexlab{}.
\newblock \showarticletitle{Query rewriting using active learning for sponsored
  search}. In \bibinfo{booktitle}{\emph{Proceedings of the 30th annual
  international ACM SIGIR conference on Research and development in information
  retrieval}}. ACM, \bibinfo{pages}{853--854}.
\newblock


\bibitem[\protect\citeauthoryear{Zhang, Zhang, Gao, Yuan, and Liu}{Zhang
  et~al\mbox{.}}{2014}]%
        {zhang2014bid}
\bibfield{author}{\bibinfo{person}{Ying Zhang}, \bibinfo{person}{Weinan Zhang},
  \bibinfo{person}{Bin Gao}, \bibinfo{person}{Xiaojie Yuan}, {and}
  \bibinfo{person}{Tie-Yan Liu}.} \bibinfo{year}{2014}\natexlab{}.
\newblock \showarticletitle{Bid keyword suggestion in sponsored search based on
  competitiveness and relevance}.
\newblock \bibinfo{journal}{\emph{Information Processing \& Management}}
  \bibinfo{volume}{50}, \bibinfo{number}{4} (\bibinfo{year}{2014}),
  \bibinfo{pages}{508--523}.
\newblock


\bibitem[\protect\citeauthoryear{Zhao, Qiu, Guan, Zhao, and He}{Zhao
  et~al\mbox{.}}{2018}]%
        {zhao2018deep}
\bibfield{author}{\bibinfo{person}{Jun Zhao}, \bibinfo{person}{Guang Qiu},
  \bibinfo{person}{Ziyu Guan}, \bibinfo{person}{Wei Zhao}, {and}
  \bibinfo{person}{Xiaofei He}.} \bibinfo{year}{2018}\natexlab{}.
\newblock \showarticletitle{Deep Reinforcement Learning for Sponsored Search
  Real-time Bidding}.
\newblock \bibinfo{journal}{\emph{arXiv preprint arXiv:1803.00259}}
  (\bibinfo{year}{2018}).
\newblock


\bibitem[\protect\citeauthoryear{Zhu, Jin, Tan, Pan, Zeng, Li, and Gai}{Zhu
  et~al\mbox{.}}{2017}]%
        {zhu2017optimized}
\bibfield{author}{\bibinfo{person}{Han Zhu}, \bibinfo{person}{Junqi Jin},
  \bibinfo{person}{Chang Tan}, \bibinfo{person}{Fei Pan},
  \bibinfo{person}{Yifan Zeng}, \bibinfo{person}{Han Li}, {and}
  \bibinfo{person}{Kun Gai}.} \bibinfo{year}{2017}\natexlab{}.
\newblock \showarticletitle{Optimized cost per click in taobao display
  advertising}. In \bibinfo{booktitle}{\emph{Proceedings of the 23rd ACM SIGKDD
  International Conference on Knowledge Discovery and Data Mining}}. ACM,
  \bibinfo{pages}{2191--2200}.
\newblock


\end{thebibliography}

% 
% If your work has an appendix, this is the place to put it.
%\appendix
%Lorem ipsum dolor sit amet, consectetur adipiscing elit. Morbi malesuada, quam in pulvinar varius, metus nunc fermentum urna, id sollicitudin purus odio sit amet enim. Aliquam ullamcorper eu ipsum vel mollis. Curabitur quis dictum nisl. Phasellus vel semper risus, et lacinia dolor. Integer ultricies commodo sem nec semper. 
%Etiam commodo feugiat nisl pulvinar pellentesque. Etiam auctor sodales ligula, non varius nibh pulvinar semper. Suspendisse nec lectus non ipsum convallis congue hendrerit vitae sapien. Donec at laoreet eros. Vivamus non purus placerat, scelerisque diam eu, cursus ante. Etiam aliquam tortor auctor efficitur mattis. 
%Nam id fermentum dui. Suspendisse sagittis tortor a nulla mollis, in pulvinar ex pretium. Sed interdum orci quis metus euismod, et sagittis enim maximus. Vestibulum gravida massa ut felis suscipit congue. Quisque mattis elit a risus ultrices commodo venenatis eget dui. Etiam sagittis eleifend elementum. 
%
%Nam interdum magna at lectus dignissim, ac dignissim lorem rhoncus. Maecenas eu arcu ac neque placerat aliquam. Nunc pulvinar massa et mattis lacinia.

\end{document}